\documentclass{bmvc2k}

\usepackage{tikz}
\usepackage{algorithm}
\usepackage[noend]{algpseudocode}
\usepackage{bm}

\usepackage{bbold}

\usepackage{times}
\usepackage{epsfig}
\usepackage{graphicx}
\usepackage{amsmath}
\usepackage{amssymb}
\usepackage{epstopdf}
\usepackage{enumitem}

\usepackage{tcolorbox}

\usepackage{caption}
\usepackage{subcaption}

\usepackage{gensymb}
\usepackage{mathtools}

\usepackage{stackengine}

\usepackage{multirow}
\usepackage{booktabs}
\usepackage{blindtext}

\usepackage{lipsum}

\newcommand\blfootnote[1]{%
  \begingroup
  \renewcommand\thefootnote{}\footnote{#1}%
  \addtocounter{footnote}{-1}%
  \endgroup
}

\title{Simple Dialogue System with AUDITED\blfootnote{$^*$ The corresponding author. {\color{black!40!blue} This paper is accepted by the British Machine Vision Conference (BMVC), 2021.}}}

\addauthor{Yusuf Tas\vspace{-0.55cm}}{}{1}
\addauthor{Piotr Koniusz$^*$}{http://users.cecs.anu.edu.au/~koniusz}{1}

\addinstitution{
$\quad\;$Data61/CSIRO, Australia \&\\
the Australian National University
}

\runninghead{Tas, Koniusz}{Simple Dialogue System with AUDITED}

\definecolor{beaublue}{rgb}{0.93, 0.97, 0.97}
\definecolor{blackish}{rgb}{0.2, 0.2, 0.2}

\definecolor{beaublue2}{rgb}{0.84, 0.9, 0.95}
\definecolor{blackish2}{rgb}{0.2, 0.2, 0.2}

\newcommand{\algblue}{black!40!blue}

\newcommand\revised[1]{{#1}}

\makeatletter
\newcommand\fs@nobottomruled{\def\@fs@cfont{\bfseries}\let\@fs@capt\floatc@ruled
  \def\@fs@pre{}
  \def\@fs@post{}
  \def\@fs@mid{\kern2pt\hrule\kern2pt}%
  \let\@fs@iftopcapt\iftrue}
\makeatother

\makeatletter
\DeclareRobustCommand\onedot{\futurelet\@let@token\bmv@onedotaux}
\def\bmv@onedotaux{\ifx\@let@token.\else.\null\fi\xspace}
\DeclareRobustCommand\onespace{\futurelet\@let@token\bmv@onespaceaux}
\def\bmv@onespaceaux{\ifx\@let@token \else \null\fi\xspace}

\def\eg{\emph{e.g}\onedot}

 \def\vs{\emph{vs}\onedot}
\def\wrt{w.r.t\onedot} 
\def\etal{\emph{et al}\onedot}
\makeatother

\renewcommand\vec[1]{\ensuremath\boldsymbol{#1}}
\renewcommand\cdots{...}

\newcommand{\vy}{\mathbf{y}}
\newcommand{\vp}{\mathbf{p}}

\newcommand{\mbr}[1]{\mathbb{R}^{#1}}

\newcommand{\vu}{\mathbf{u}}

\newcommand{\vphi}{\boldsymbol{\phi}}
\newcommand{\vpsi}{\boldsymbol{\psi}}

\newcommand{\mPsi}{\vec{\Psi}}

\DeclareMathOperator*{\maxx}{max}

\newcommand{\mLambda}{\bm{\lambda}}

\def\eg{\emph{e.g.}}

\newcommand{\mTheta}{\boldsymbol{\Theta}}

\newcommand{\vh}{\boldsymbol{h}}
\newcommand{\vmu}{\boldsymbol{\mu}}

\newcommand{\stkout}[1]{{\ifmmode\text{\sout{\ensuremath{#1}}}\else\sout{#1}\fi}}

\renewcommand{\comment}[1]{}

\begin{document}

\maketitle

\begin{abstract}
We devise a multimodal conversation system for dialogue utterances composed of text, image or both modalities. We leverage Auxiliary UnsuperviseD vIsual and TExtual Data (AUDITED). To improve the performance of text-based task, we utilize translations of target sentences from English to French to form the assisted supervision. For the image-based task, we employ the DeepFashion  dataset in which we seek nearest neighbor images of positive and negative target images of the MMD data. These nearest neighbors form the nearest neighbor embedding providing an external context for target images. We form two methods to create neighbor embedding vectors, namely Neighbor Embedding by Hard Assignment (NEHA) and Neighbor Embedding by Soft Assignment (NESA) which generate context subspaces per target image. Subsequently, these subspaces are learnt by our pipeline as a context for the target data. We also propose a discriminator which switches between the image- and text-based tasks. We show improvements over  baselines on the large-scale Multimodal Dialogue Dataset (MMD) and SIMMC.
\end{abstract}

\section{Introduction}
\label{sec:intro}

\begin{figure}
\centering
\begin{subfigure}[t]{0.77\linewidth}
\centering\includegraphics[width=1\linewidth]{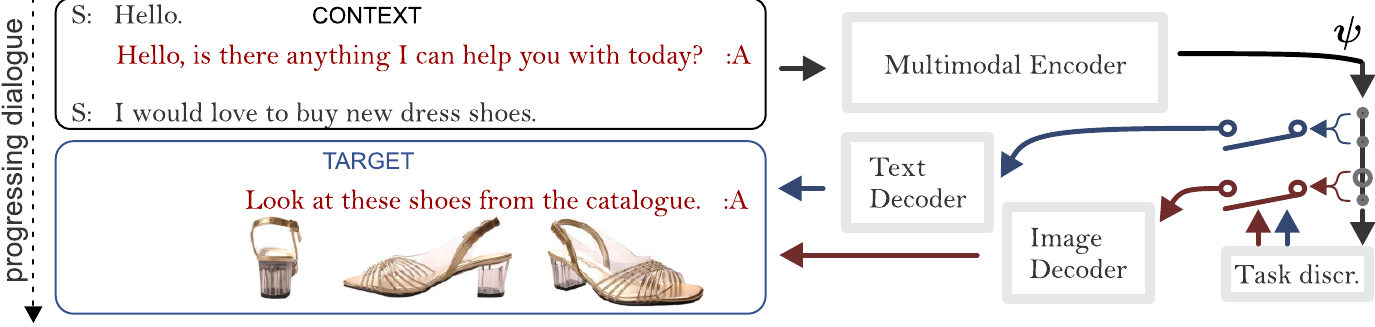}
\vspace{-0.5cm}
\caption{\label{fig:dialogue}}
\end{subfigure}
\hspace{0.2cm}
%
\begin{subfigure}[t]{0.19\linewidth}
\centering\includegraphics[width=1\linewidth]{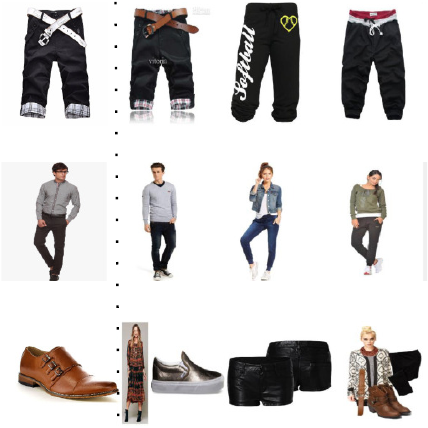}
\vspace{-0.5cm}
\caption{\label{fig:nnimages}}
\end{subfigure}
\vspace{0.3cm}
\caption{\revised{Our pipeline includes the Multimodal Encoder, Text Decoder, Image Decoder (Feature Matching Head) and the Task Discriminator (Fig. \ref{fig:dialogue}). The MMD dataset contains dialogues between shoppers ({\em S}) and retail agents ({\em A}) which progress in time. Dialogues are split by the sliding window (default protocol) to form the input ({\em CONTEXT}) fed to the Multimodal Encoder. The output ({\em TARGET}) may contain text, images, or both modalities, which are imposed via dedicated losses on the Text and/or Image Decoders.  The switches indicate that one half of the Context Descriptor $\vpsi$ may be passed to the Text Decoder and the other half to the Image Decoder depending on the Task Discriminator. The details of Multimodal Encoder, Text Decoder and Image Decoder  are shown in Figures \ref{fig:enc1}, \ref{fig:mm_decoder_t} and \ref{fig:pipeline}, respectively. 
Figure \ref{fig:nnimages} shows 3 nearest neighbors (columns 2--4) retrieved (decreasing similarity order) from DeepFashion \cite{liu2016deepfashion} for query samples (column 1) from the MMD dataset \cite{saha2018towards}. Feature descriptors were encoded by ResNet-50, the approximate nearest neighbor search was performed by the FAISS library \cite{johnson2019billion}. Such images form an external context for target images.}}
\label{fig:dialogue_all}
\vspace{-0.3cm}
\end{figure}

Deep learning is popular in many areas \eg, object detection \cite{girshick2014rich}, speech recognition \cite{graves2013speech}, image super-resolution \cite{dong2015image}, text and natural language processing \cite{devlin2018bert}, domain adaptation \cite{koniusz2017domain,Koniusz2018Museum,action_da}, few-shot learning \cite{sosn,Zhang_2020_ACCV,arl,kon_tpami2020a}, and even arts recognition \cite{zhang2018artwork,Koniusz2018Museum}. 
%
Realistic problems such as Visual Question Answering (VQA) are often multimodal. 
Image Captioning (IC) \cite{xu2015show} learns from  text and images to generate image captions. VQA \cite{zeng2017leveraging} answers questions about a video by leveraging the spatio-temporal visual data and the accompanying text. 
Multimodal conversation systems use text and images used together as chat bots \cite{ram2018conversational}, autonomous retail agents \cite{saha2018towards} and task-specific dialogue systems \cite{wen2016network}. Saha \etal \cite{saha2018towards} introduced one of the largest multimodal conversation datasets called Multimodal Dialogue (MMD) dataset, containing over 150K shopper-retail agent dialogues. 
Figure \ref{fig:dialogue} shows dialogues of shoppers asking about/referring to items or asking for items from a given image. MMD contains the image- and text-based tasks. In the image-based task, the model has to retrieve/rank the correct image from given positive and negative images in response to the multimodal context. The text-based task  predicts the agent's response within the  context. 

In this paper, we go beyond separate protocols of Saha \etal \cite{saha2018towards} by introducing a discriminator whose role is to learn/predict an appropriate task. 
As a limited number of utterances contain images, we leverage external visual and textual knowledge via the so-called assisted supervision. Figure \ref{fig:dialogue} shows our pipeline. Our contributions are listed below:

\renewcommand{\labelenumi}{\roman{enumi}.}
\vspace{-0.5cm}\hspace{-2cm}
\begin{enumerate}[leftmargin=0.5cm]
\vspace{-0.1cm}
\item We propose a novel assisted supervision to create a context for target images and thus implicitly incorporate more images in unsupervised manner into the learning process of image-based task. The DeepFashion dataset \cite{liu2016deepfashion} is used to search for closest matching images to given  positive and negative target images. Through the perspective of sampling the natural manifold of images, we capture context images for target images.
\vspace{-0.3cm}
\item We design two embeddings for neighbor images: Neighbor Embedding by Hard Assignment (NEHA) and Neighbor Embedding by Soft Assignment (NESA). NEHA  retrieves $\eta$ nearest neighbors for positive/negative target images to encode them into subspace descriptors by SVD. NESA also reweights the contribution of each context image by the membership probability in a GMM-like model \cite{pk_cviu,pk_hop} spanned on target images.
\vspace{-0.3cm}
\item For the text-based task, we propose   an assisted supervision that uses translation decoders to generate predictions of text 
in multiple languages to  learn a universal representation of conversations by limiting ambiguities of a single language model 
\cite{cog_ben}.  
\vspace{-0.3cm}
\item Finally, we introduce a discriminator whose role is to combine image- and text-based tasks by learning to predict an appropriate task in response given the multimodal context.
\end{enumerate}

\revised{The above strategy of leveraging unsupervised data can be seen as capturing the variance of linguistic and visual data to help the network capture how each utterance may vary.}

\section{Related Work}
\label{sec:related_work}

\vspace{-0.1cm}
Below we describe popular dialogue systems, and detail the Multimodal Hierarchical Encoder Decoder (M-HRED) \cite{saha2018towards} on which we build. 

\vspace{0.1cm}
\noindent\textbf{\fontsize{11}{11}\selectfont Conversation Systems.} Early conversation systems \cite{banchs2012movie, ameixa2014luke} use scripts and subtitles for retrieval of responses in a dialogue. Ritter \etal \cite{ritter2010unsupervised} uses generative probabilistic models for  conversations on blogging websites. VQA approaches \cite{antol2015vqa, wu2018joint} answer questions about images. Approaches \cite{das2017visual, kottur2019clevr} tackle visual dialogs about individual images. 
Approach \cite{thomason2020vision} focuses on the visual dialog navigation. Bhattacharya \etal \cite{bhattacharya2019multimodal} retrieves images via textual queries.  FashionIQ \cite{guo2019fashion} is concerned with the NLP-based image retrieval. 

Recent dialogue systems use an RNN encoder-decoder \cite{sordoni2015neural, shang2015neural}. Hierarchical Recurrent Encoder-Decoder \cite{serban2015building} uses a two-level RNN to create a context-aware conversation system. Approach \cite{saha2018towards} predicts answers of a shopping assistant from natural conversations of the large scale MMD dataset, which we use. Below, we describe and build on models \cite{serban2016building,serban2017hierarchical}. 

\begin{figure}[t]
%
\begin{subfigure}[b]{0.495\linewidth}
\centering\includegraphics[width=5.0cm]{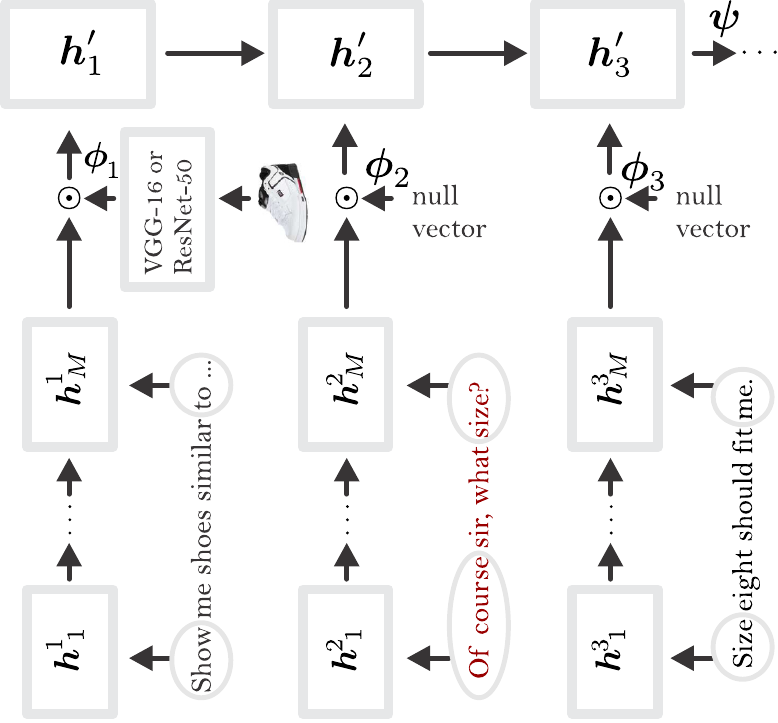}
\vspace{-0.1cm}
\caption{\label{fig:enc1}}
\vspace{0.2cm}
\end{subfigure}
\begin{subfigure}[b]{0.495\linewidth}
\centering\includegraphics[width=5.75cm]{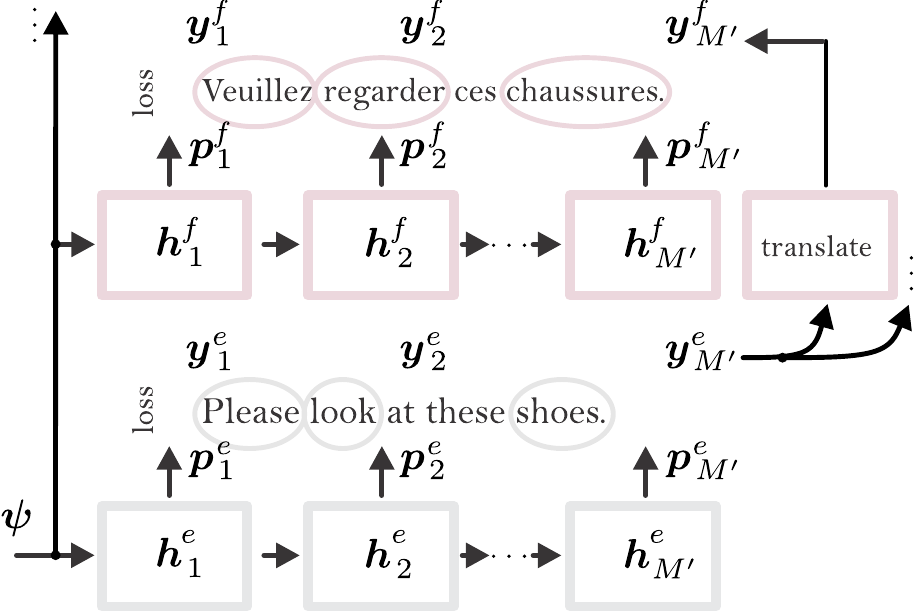}
\vspace{-0.1cm}
\caption{\label{fig:mm_decoder_t}}
\vspace{0.2cm}
\end{subfigure}
%
%
\caption{\revised{In Multimodal Encoder, shown in Fig. \ref{fig:enc1}, the text is processed by a first-level GRU  while images are encoded by ResNet-50 to obtain compact embeddings. 
We concatenate text and avg-pooled image representations (if image is not present, we use a null vector) by $\odot$ into utterance descriptors $\vphi_1,\cdots,\vphi_3$ and process them with a second-level GRU to produce the Context Descriptor $\vpsi$, which we pass it to
the Text Decoder with the assisted supervision in Figure \ref{fig:mm_decoder_t}. The text is translated from English (ground truth) into French (and other languages). The losses (per language) encourage the network to absorb syntactic differences which implicitly helps capture the true dynamics of the dialogue better. Standard Text Encoder \cite{saha2018towards} consists  of the gray blocks while pink blocks form our assisted supervision.}
}
\label{fig:mm_encoders}
\vspace{-0.2cm}
\end{figure}


\vspace{0.1cm}
\noindent\textbf{\fontsize{11}{11}\selectfont Multimodal-Hierarchical Encoder Decoder.} 
\label{sec:hed}
M-HRED \cite{saha2018towards} is an extension of Hierarchical Recurrent Encoder Decoder (HRED) models \cite{serban2016building,serban2017hierarchical}. HRED  consists of two different levels of Recurrent Neural Network (RNN) \cite{mikolov2010recurrent} combined together, which represent an encoder which captures the so-called word and sentence context, respectively. The first RNN in HRED model learns to generate the next word in a given sentence by using the word context. The second RNN takes the final representation of a given sentence to generate the  representation of next sentence by using the sentence context. An RNN decoder receives a sentence-level representation to decode it and generate a full sentence. Moreover, M-HRED and HRED  use the interconnected encoder and decoder but M-HRED also uses  images. 

\vspace{0.1cm}
\noindent{\textbf{Multimodal Encoder (ME).}} ME receives a sequence of $N$ utterances (so-called context) to produce the Context Descriptor via GRU \cite{chung2014empirical}. An utterance contains a sentence, image or both modalities. Images are encoded by VGG-16 \cite{simonyan2014very} (4096 ch. of the last FC layer \cite{saha2018towards}). 

\begin{figure*}[t]
\centering
\centering\includegraphics[width=1.0\textwidth]{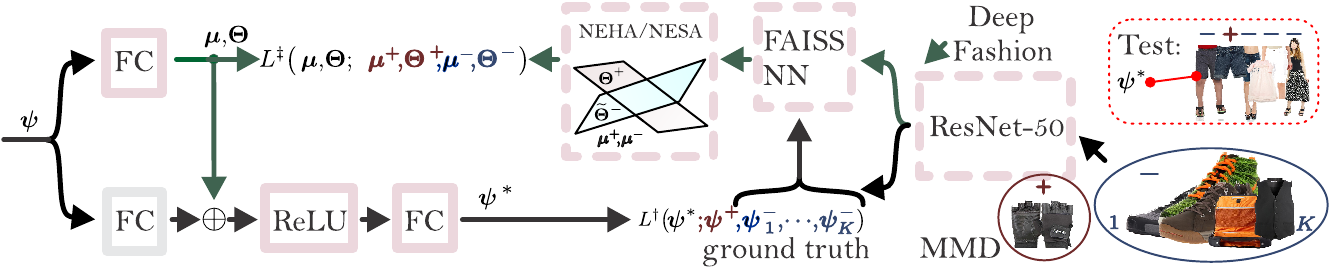}
%
\vspace{0.001cm}
\caption{Our Image Decoder a.k.a. Feature Matching Head consists of the main stream (FC$\rightarrow$ReLU$\rightarrow$FC) whose role is to take Context Descriptors $\vpsi$ and produce visual features $\vpsi^*$ that are combined with loss $L^\dagger$ in Eq. \eqref{eq:image1}. The traditional head (older method) contains one FC layer (gray block). For each ground truth target positive and negative  image descriptors $\vpsi^+$ and $\vpsi_1^-,\cdots,\vpsi_K^-$ from the MMD dataset (encoded by ResNet-50), we find $\eta$ and $\eta K$ approximate nearest neighbor image descriptor from DeepFashion \cite{liu2016deepfashion} with the FAISS library \cite{johnson2019billion}. Then we create positive and negative mean descriptors $\vmu^+$ and $\vmu^-$ as well as subspaces $\mTheta^+$ and $\mTheta^-$ with NEHA or NESA step. \revised{They capture the mean and variability of positive and negative images.} The role of another FC layer is to learn  positive visual context representations $(\vmu,\mTheta)$ via the assisted supervision loss $L^\ddagger$ in Eq. \eqref{eq:assist} which attracts $(\vmu,\mTheta)$ towards $(\vmu^+,\mTheta^+)$ and repels it from $(\vmu^-,\mTheta^-)$. Finally,  $(\vmu,\mTheta)$ are combined with the main stream via a residual link (operator $\oplus$). Blocks with dashed borders/losses are not used during testing. \revised{During testing, $\vpsi^*$ is matched against test images of an utterance. In the above example, the correct ground truth is ranked as second (R@1 fails but R@2 succeeds).}}
\label{fig:pipeline}
\vspace{-0.3cm}
\end{figure*}

\revised{Multimodal Utterance Encoder (MUE) in Fig. \ref{fig:enc1} consists of two levels of GRU \cite{chung2014empirical}. The first-level GRU (bottom) contains hidden states $\vh^n_1,\cdots,\vh^n_M$, where $M$ is the maximum number of input words per utterance, each word is one-hot encoded with a discrete vocabulary of size $V\!=\!7457$, $n\!=\!1,\cdots,N$ and $N$ is the context size \eg, $N\!=\!3$ utterances. The first-level GRU and ResNet-50 encode words and images, respectively. The last state and the output of ResNet-50 are concatenated by $\odot$ into $\vphi$ and padded with zeros if image or text is missing. Encoded utterances are  passed to the Context Encoder (CE), a second-level GRU, with hidden states $\vh'_1,\cdots,\vh'_{N'}$ shown in Fig. \ref{fig:enc1} (top) to obtain a Context Descriptor $\vpsi$ per context. Fig. \ref{fig:dialogue} shows examples of context and target utterances. %
We use encoder networks from the M-HRED model \cite{saha2018towards} (results in the same testbed, based on ResNet-50).}

\vspace{0.1cm}
\noindent{\textbf{Multimodal Decoder (MD).}} MD receives the Context Descriptor $\vpsi$ from CE. 
For the text-based task, the target is a sentence. A GRU decoder \cite{serban2016building} with hidden states $\vh^e_1,\cdots,\vh^e_{M'}$ generates the target sentence word-by-word, starting with the start-of-sentence and ending with end-of-sentence token. Given the target ground truth sentence with one-hot representation of words $\vy^e$ and the final output predictions $\vp^e$ from the model ($e$ indicates English), the combined Multimodal Encoder Decoder is trained via the cross-entropy loss. \revised{At the test time, the quality of generated utterance is evaluated against target ground truth sentences via so-called BLEU and NIST metrics \cite{saha2018towards}.} Figure \ref{fig:mm_decoder_t} shows our extended Text Decoder (pink plus gray blocks) and the baseline Text Decoder (gray block) \cite{saha2018towards}. 
\revised{The image-based task is the ranking-based task. Given a positive target image, and $K$ negative images, Context Descriptor $\vpsi$ is ranked against these positive and negative images at the test time.} During training,  M-HRED  uses the cosine similarity and the hinge loss:

\vspace{-0.7cm}
\begin{align}
&\!\!\!\!L^\dagger\left(\vpsi^*;\vpsi^{+}\!,\vpsi_1^{-}\!,\cdots,\vpsi_K^{-}\right) \!=\!
\text{max}\Big(0, 1 \!-\! {\vpsi^{*}}^{T}\Big(\vpsi^{+} \!-\! \frac{1}{K}\sum\limits_{k=1}^K\vpsi_k^{-}\Big)\Big),
\label{eq:image1}
\end{align}

\vspace{-0.3cm}
\noindent where $\vpsi^*\!\!\in\!\mbr{4096}$ is a feature vector obtained by passing the Context Descriptor $\vpsi\!\in\!\mbr{1024}$ from CE via an FC layer, and $\vpsi^{+}\!\!\in\!\mbr{4096}$ and $\vpsi^{-}\!\!\in\!\mbr{4096}$ correspond to  image descriptors (VGG-16) for the positive and negative ground truth images, resp. The Hinge loss encourages $\vpsi^*$ to be close to $\vpsi^{+}$ and away from $\vpsi^{-}$. $L$ is minimized \wrt network parameters.

\vspace{0.1cm}
\noindent\textbf{\fontsize{11}{11}\selectfont Self-supervised Learning.} 
Pretext tasks such as sampling and predicting patch locations (left, right, top left, top right), rotations (0\degree, 90\degree, 180\degree, 270\degree) or other transformations are popular in self-supervision \cite{doersch2015unsupervised,dosovitskiy2015discriminative,gidaris2018unsupervised,zhang2020few,arl}. 
%
Note  self-supervision by mutual information estimation \cite{hjelm2018learning}, egomotion prediction \cite{agrawal2015learning}, and   multi-task self-supervised learning \cite{doersch2017multi}. One


\algblock{while}{endwhile}

\algblock[TryCatchFinally]{try}{endtry}
\algcblockdefx[TryCatchFinally]{TryCatchFinally}{catch}{endtry}
	[1]{\textbf{except}#1}{}
\algcblockdefx[TryCatchFinally]{TryCatchFinally}{elsee}{endtry}
	[1]{\textbf{else}#1}{}
	
\algtext*{endwhile}

\algtext*{endtry}

\algblockdefx{ifff}{endifff}
	[1]{\textbf{if}#1}{}
\algtext*{endifff}

\algblockdefx{forr}{endforr}
	[1]{\textbf{for}#1}{}
\algtext*{endforr}

\algblockdefx{forrr}{endforrr}
	[2]{{\color{#2}\textbf{for}#1}}{}
\algtext*{endforrr}

\noindent\begin{minipage}[!t]{0.495\textwidth}
%
%
{\fontsize{7}{7}\selectfont
\begin{tcolorbox}[width=1.0\linewidth, colframe=blackish,colback=beaublue, boxsep=0mm, arc=3mm, left=1mm, right=1mm, right=1mm, top=1mm, bottom=1mm]
{\bf Input:} $\eta'\!\!\leq\!\eta$, $K$, $L$\\
$\vpsi_1^+$, $\mPsi^-\!\!\equiv\!\{\vpsi_1^-,\cdots,\vpsi_K^-\}\gets$ ground truth positive and negative target descriptors from MMD,\\
$\{\vpsi'_1,\cdots,\vpsi'_L\}\gets$ unsupervised feature descriptors from  DeepFashion \cite{liu2016deepfashion}.
\begin{algorithmic}[1]
	\State $({\vpsi'}_1^+,\cdots,{\vpsi'}_\eta^+)\!=\!\text{FAISS\_NN}\left(\vpsi^+, \eta;\;\{{\vpsi'}_1,\cdots,{\vpsi'}_L\}  \right)$
	\forrr{ $n\!=\!1,\cdots,\eta$:}{\algblue}
	\State{\color{\algblue} ${\vpsi'}_n^+ \gets s^+({\vpsi'}_n^+,\vpsi^+\!,\mPsi^-)\cdot{\vpsi'}_n^+$ }
	\endforrr
	\State{$\vmu^+\!=\!\frac{1}{\eta}\sum\limits_{n=1}^\eta {\vpsi'}_n^+$}
	\State{$(\mTheta^+,\mLambda^+)\!=\!\text{SVD}({\vpsi'}_1^+\!-\!\vmu^+,\cdots,{\vpsi'}_\eta^+\!-\!\vmu^+;\;\eta')$}
		\forr{ $k\!=\!1,\cdots,K$:}
		\State $({\vpsi'}_{1k}^-,\cdots,{\vpsi'}_{\eta k}^-)\!=$
		\State $\qquad\text{FAISS\_NN}\left(\vpsi_k^-, \eta;\;\{{\vpsi'}_1,\cdots,{\vpsi'}_L\}  \right)$
		\endforr
	\forrr{ $k\!=\!1,\cdots,K$:}{\algblue}
	\forrr{ $n\!=\!1,\cdots,\eta$:}{\algblue}
	\State{\color{\algblue} ${\vpsi'}_{nk}^- \gets s^+({\vpsi'}_{nk}^-,\vpsi^+\!,\mPsi^-)\!\cdot\!{\vpsi'}_{nk}^-$}
	\endforrr
	\endforrr
	\State{$\vmu^-\!=\!\frac{1}{\eta K}\sum\limits_{n=1}^\eta\sum\limits_{k=1}^K {\vpsi'}_{nk}^-$}
	\State{$(\mTheta^-,\mLambda^-)\!=\!\text{SVD}({\vpsi'}_1^-\!-\!\vmu^-,\cdots,{\vpsi'}_{\eta K}^-\!-\!\vmu^-;\;\eta')$}
\end{algorithmic}
{\bf Output:} $(\vmu^+,\mTheta^+)$ and $(\vmu^-,\mTheta^-)$
\end{tcolorbox}
\vspace{0.2cm}
\captionof{algorithm}{Neighbor Embedding by Hard Assignment (black color). {\color{\algblue}Neighbor Embedding by Soft Assignment (black/blue colors)}.}
\label{alg:neha}
}
\vspace{0.2cm}
%
\end{minipage}
\hspace{0.3cm}
\begin{minipage}[!t]{0.495\textwidth}
\captionsetup{type=figure}
%
\begin{subfigure}[b]{0.49\linewidth}
\centering\includegraphics[width=3.25cm]{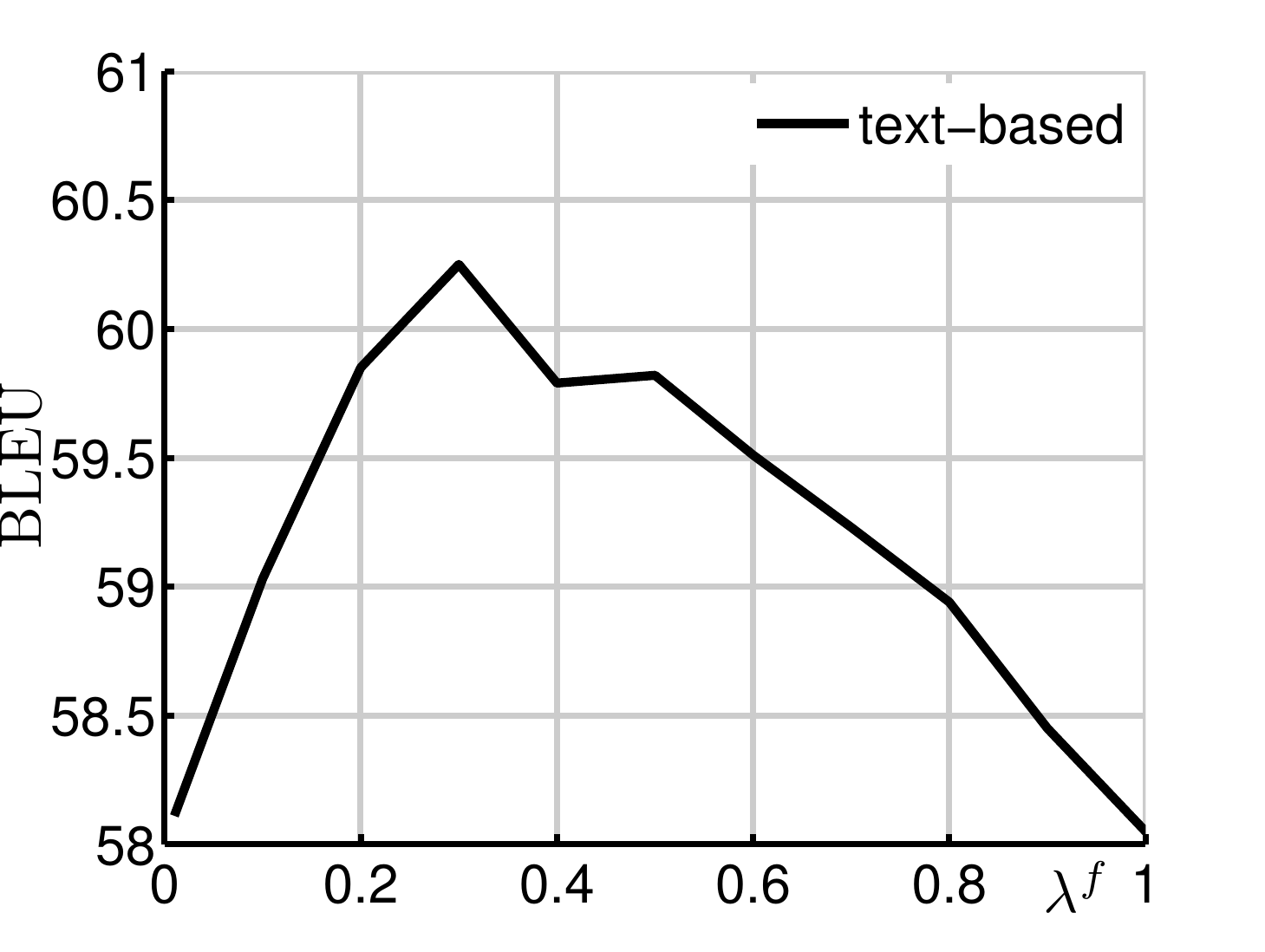}
\vspace{-0.6cm}
\caption{\label{fig:cv1}}
\vspace{0.5cm}
\end{subfigure}
\begin{subfigure}[b]{0.49\linewidth}
\centering\includegraphics[width=3.25cm]{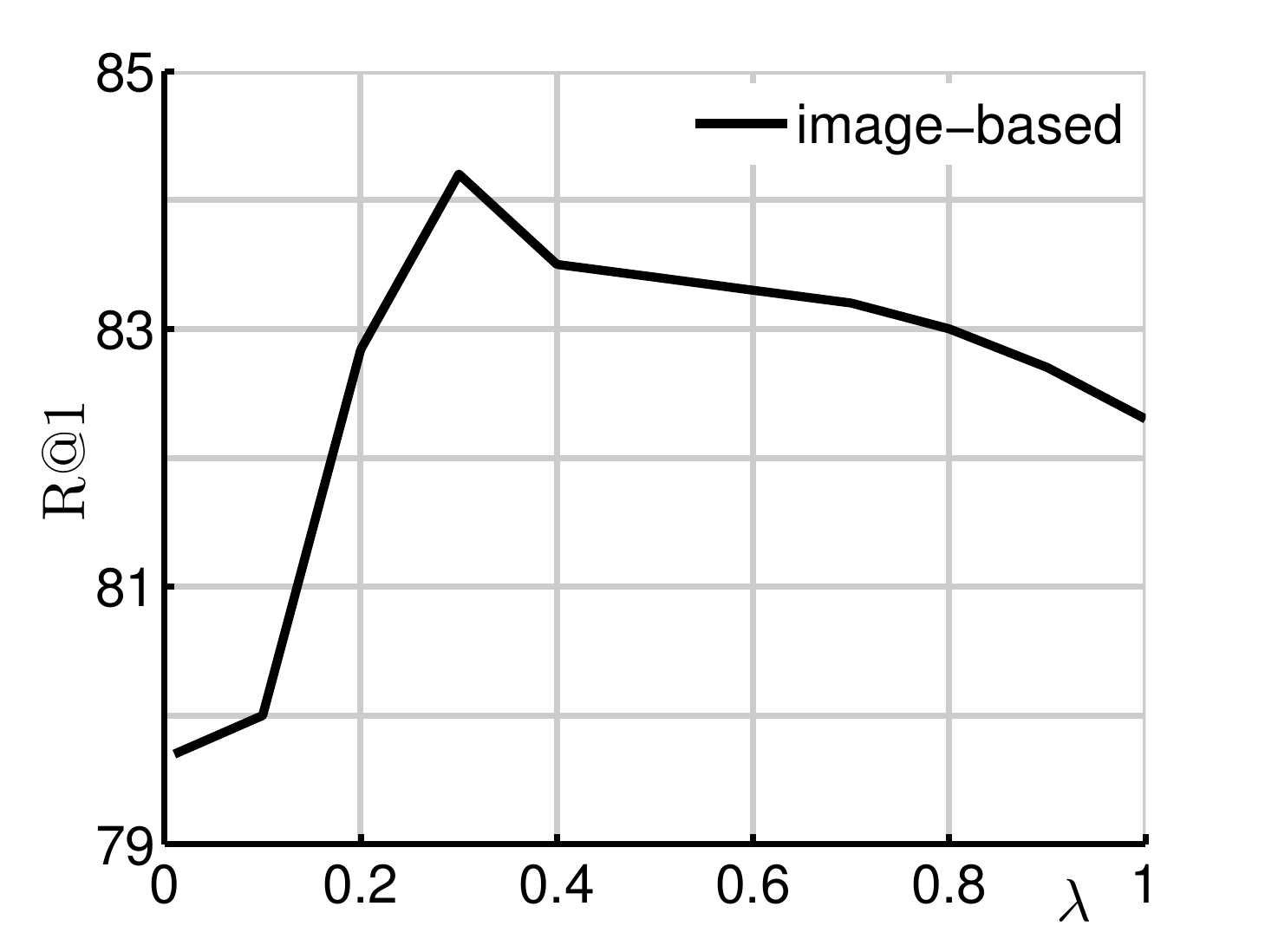}
\vspace{-0.6cm}
\caption{\label{fig:cv2}}
\vspace{0.5cm}
\end{subfigure}\\
\begin{subfigure}[b]{0.49\linewidth}
\centering\includegraphics[width=3.25cm,trim=20 190 -10 320]{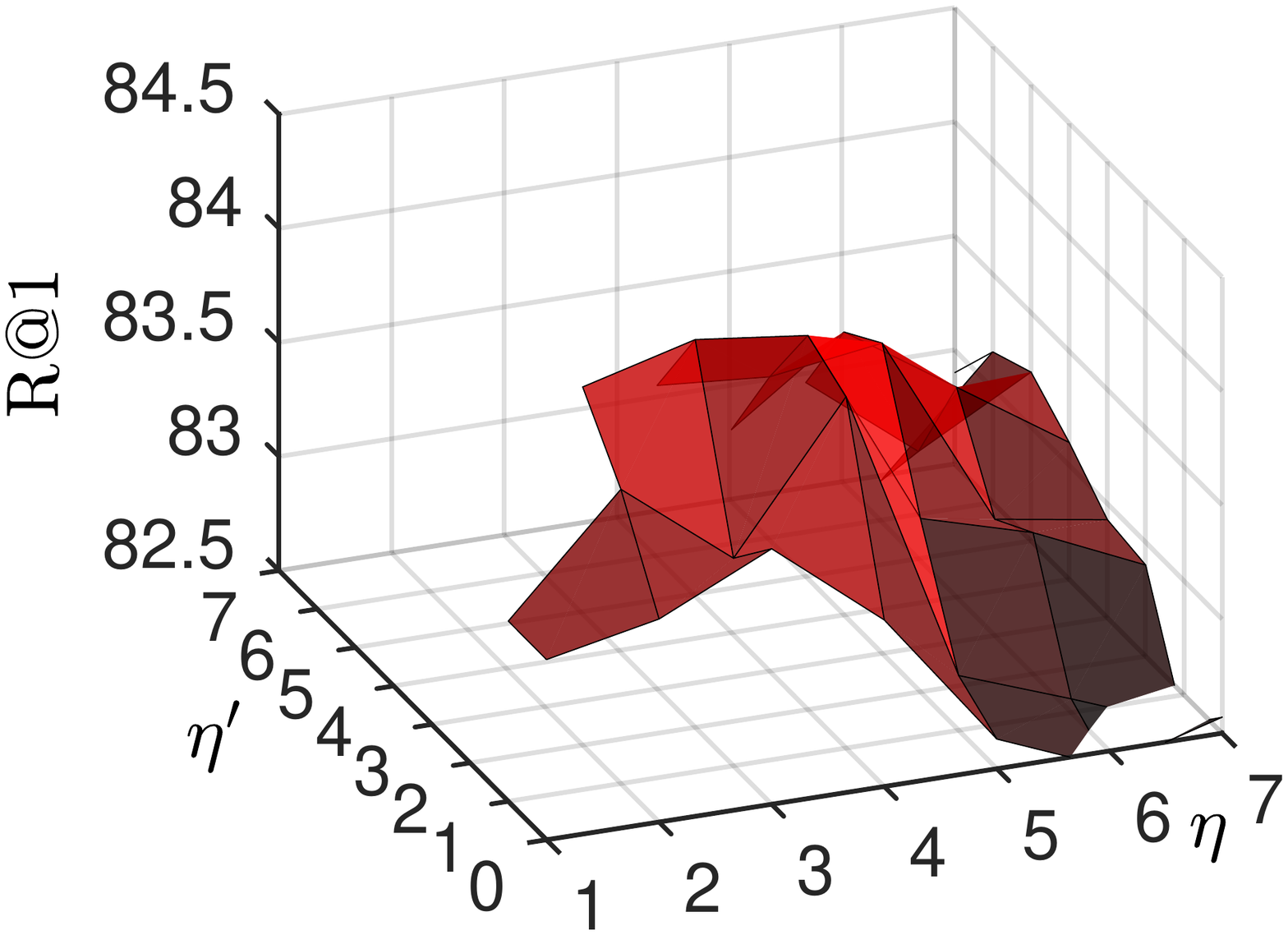}
\vspace{-0.6cm}
\caption{\label{fig:cv3}}
\end{subfigure}
\begin{subfigure}[b]{0.49\linewidth}
\centering\includegraphics[width=3.25cm,trim=20 190 -10 320]{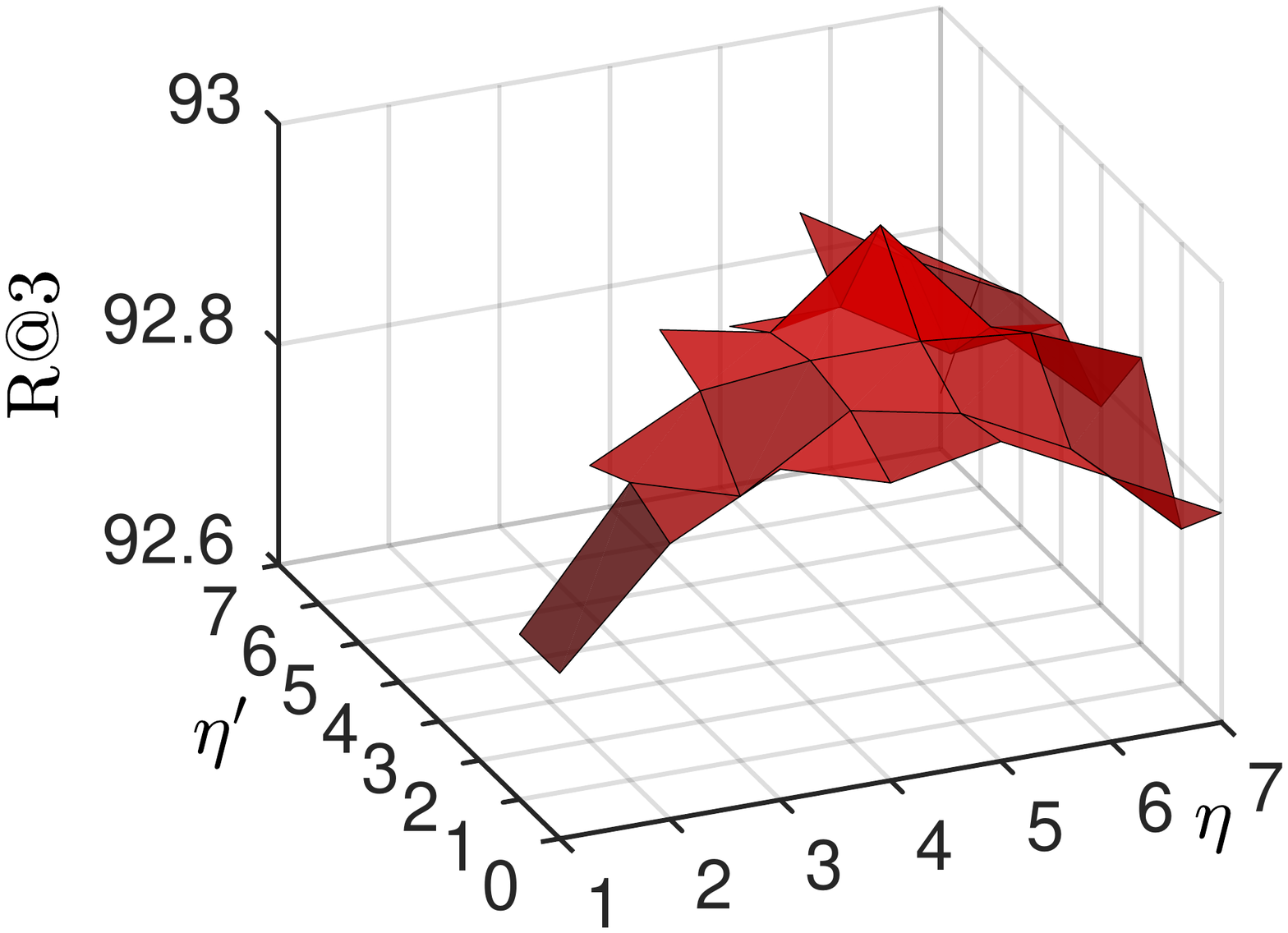}
\vspace{-0.6cm}
\caption{\label{fig:cv4}}
\end{subfigure}
%
{Fig. \ref{fig:cv1} \& \ref{fig:cv2} are cross-validation results \wrt $\lambda^f$ and $\lambda$ for the text- and image-based tasks. Fig. \ref{fig:cv3} \& \ref{fig:cv4} are cross-val. results (R$@1$ \& R$@3$ metric) 
\wrt $\eta'$ and $\eta$ for the image-based task. If $\eta'\!=\!0$, we use $\mu^-$ and $\mu^+$ only.}
\vspace{0.2cm}
\end{minipage}

\noindent may use Bag-of-Words on hand-crafted descriptors for an alignment task \cite{i3d_halluc,i3d_halluc2}, or form positive and negative sampling for a contrastive learning strategy \cite{ssgc,refine,coles}. 
 GAN-based pipelines \cite{GAN,ShiriYPHK19,fatima_ijcv} also perform self-supervision by generator-discriminator competition.

\revised{
\vspace{0.1cm}
\noindent\textbf{\fontsize{11}{11}\selectfont Motivation from Cognitive Psychology.} 
For the text-based task, we use a translating network \cite{ott2018scaling} and decoders to predict target responses in several languages. This limits the quantization noise resulting from the single language syntactic thus helping capture universal concepts better. 
 Cognitive psychology notes that multilingual babies exhibit better attention and conflict management, and adjust to new rules  quicker than monolingual babies  \cite{cog_ben}.

For the image-related task, we retrieve  the $\eta$ and $\eta K$ nearest neighbors from  the DeepFashion \cite{liu2016deepfashion} dataset for positive and negative  target images to form subspace descriptors  which represent the learning context of target images, and form the manifold of fashion images. From the psychological point of view, our approach is motivated by knowledge transfer,  which is `{\em the dependency of human conduct, learning or performance on prior experience'} a.k.a. `{\em transfer of particle}' \cite{woodworth_particle}. Notice that pre-training our visual task on the DeepFashion is impossible as DeepFashion dataset is not organised in the form of dialogue.

In conclusion, providing multiple translations and multiple positive images (subspaces are second-order statistics) helps our pipeline capture better the innate variance of data.

%
}


\vspace{-0.2cm}
\section{Our Approach}
\label{sec:our_approach}
\vspace{-0.1cm}


%
\vspace{0.1cm}
\noindent{\textbf{\fontsize{11}{11}\selectfont Notations.}} Bold lowercase symbols are vectors \eg, $\vmu,\vphi,\vpsi$. Regular lowercase/uppercase symbols are scalars \eg, $\eta, K, N$. Bold uppercase symbols are matrices or sets of parameters \eg, $\mTheta$. Symbols $\odot$ and $\oplus$ are the vector concatenation \& summation (residual link).

%
%


\vspace{0.1cm}
\noindent\textbf{\fontsize{11}{11}\selectfont Pipeline.}\label{sec:pipe} Our pipeline in Figure \ref{fig:dialogue} follows the baseline model \cite{saha2018towards} in that we use the Multimodal Encoder, Text Decoder and Image Decoder (Feature Matching Head only). \revised{The Multimodal Encoder receives the context, a collection of $N\!=\!3$ utterances which are snippets of dialogues between a shopper and a retail agent obtained by a sliding window, a standard protocol on the MMD dataset of retail dialogues. The context window may contain text, image, or both modalities. The Multimodal Encoder  takes $N$ utterances, based on a discrete vocabulary of size $V$, and ResNet-50 encoded images to produce the Context Descriptor which is fed to the Text Decoder and Image Decoder, whose roles are to predict a target ground truth text responses (within discrete vocabulary space) and/or generate ResNet-50 image features to retrieve a visual recommendation from the MMD (or SIMMC) dataset (also encoded with ResNet-50). As the baseline model \cite{saha2018towards} is formulated as two separate tasks, it requires ground truth test labels about the type of output task to perform. In contrast, we introduce the Task Discriminator (the pink box in Figure \ref{fig:dialogue} which resolves this issue.  
To improve predictions, our Text and Image Decoders use the assisted supervision by leveraging the knowledge from the DeepFashion \cite{liu2016deepfashion} dataset and the translation model \cite{ott2018scaling} in an unsupervised way.  
Section \ref{sec:hed} details the Multimodal Encoder. Below we detail our decoders.}

\vspace{0.1cm}
\noindent\textbf{\fontsize{11}{11}\selectfont Image-based Task.}\label{sec:im} Figure \ref{fig:pipeline} shows our Feature Matching Head (Image Decoder). The image-based task finds the closest match between a predicted image descriptor and one positive and $K$ negative ground truth descriptor candidates per target utterance. The image-based task uses two losses, the standard loss $L^\dagger$ given by Eq. \eqref{eq:image1} and our assisted supervision loss:
\vspace{-0.5cm}
\begin{align}
&\!\!\!\!L^\ddagger\left(\vmu,\mTheta;\;\vmu^{+}\!,\mTheta^{+}\!,\vmu^{-}\!,\mTheta^{-}\right)\!=\!
\text{max}\Big(0, 1 \!-\! \vmu^T(\vmu^{+} \!-\! \vmu^{-}) \!-\! \sum\limits_{n=1}^{\eta'}\vu_n^T(\vu_n^{+}\!-\!\vu_n^{-})  \Big),
\label{eq:assist}
\end{align}

\vspace{-0.4cm}
\noindent where $\vpsi^*\!\!\in\!\mbr{D}$ is a feature vector of size $D\!=\!2048$ obtained by passing the Context Descriptor $\vpsi\!\in\!\mbr{1024}$ from CE via an FC layer. Moreover, $\vmu\!\in\!\mbr{D}$ and $\mTheta\!\equiv\![\vu_1,\cdots,\vu_{\eta'}]\!\in\!\mbr{D\times\eta'}$ are the context feature vectors generated by an FC layer, indicated in Figure \ref{fig:pipeline}, which are encouraged by a Hinge loss to approach the mean $\vmu^+\!\!\in\!\mbr{D}$ and eigenvectors $\mTheta^+\!\!\equiv\![\vu^+_1,\cdots,\vu^+_{\eta'}]\!\in\!\mbr{D\times\eta'}$ and stay repelled from the mean $\vmu^-\!\!\in\!\mbr{D}$ and eigenvectors $\mTheta^-\!\!\equiv\![\vu^-_1,\cdots,\vu^-_{\eta'}]\!\in\!\mbr{D\times\eta'}$. Visual Feature Descriptors (VFD) $(\vmu^+,\mTheta^+)$ and $(\vmu^-,\mTheta^-)$ represent the positive and negative context for the ground truth positive and negative target descriptors $\vpsi^+\!\!\in\!\mbr{D}$ and $\vpsi_1^-,\cdots,\vpsi_K^-\!\!\in\!\mbr{D}$ obtained from ResNet-50. Below we explain Neighbor Embedding by Hard Assignment (NEHA) and Neighbor Embedding by Soft Assignment (NESA) which produce VFDs.

\vspace{0.1cm}
\noindent{\textbf{NEHA}} is obtained by applying SVD to $\eta$ and $\eta K$ nearest neighbors ${\vpsi'}_1^+,\cdots,{\vpsi'}_\eta^+\!\!\in\!\mbr{D}$ and ${\vpsi'}_{11}^-,\cdots,{\vpsi'}_{\eta K}^-\!\!\in\!\mbr{D}$ found among images of DeepFashion \cite{liu2016deepfashion} dataset encoded by ResNet-50, represented by $L$ feature descriptors  $\vpsi'_1,\cdots,\vpsi'_L$. The search is performed by FAISS \cite{johnson2019billion}, an extremely efficient approximate nearest neighbor search library, by searching feature descriptors of DeepFashion against the ground truth positive/negative target descriptors $\vpsi^+$ and $\vpsi_1^-,\cdots,\vpsi_K^-$ from the MMD dataset, respectively. Figure \ref{fig:nnimages} shows the quality of matching images from DeepFashion against ground truth images from  MMD.

Algorithm \ref{alg:neha} shows  steps of NEHA.  $\text{FAISS\_NN}(\vpsi, \eta;\;\{{\vpsi'}_1,\cdots,{\vpsi'}_L\})$ denotes the FAISS search which retrieves $\eta$ approximate nearest neighbors of $\vpsi$ from $\{{\vpsi'}_1,\cdots,{\vpsi'}_L\}$. Moreover, $\text{SVD}({\vpsi}_1,\cdots,{\vpsi}_\eta;\;\eta')$ returns $\eta'\!\leq\!\eta$ leading eigenvectors and eigenvalues $(\mTheta,\mLambda)$. 
We note that NEHA does not take into account the effect of decreasing similarity between the ground truth positive/negative target descriptors and searched feature descriptors of DeepFashion as one progresses over consecutive $1,\cdots,\eta$ nearest neighbors. Thus, Visual Feature Descriptors $(\vmu^+,\mTheta^+)$ and $(\vmu^-,\mTheta^-)$ may provide gradually worsening visual context for target descriptors of MMD. To his end, we introduce an improved strategy below.

\vspace{0.1cm}
\noindent{\textbf{NESA}}  follows  NEHA but it uses reweighting 
by so-called Soft Assignment applied prior to SVD steps. We use the two weighting functions for positive ${\vpsi'}^+$ and negative ${\vpsi'}^-$:
\vspace{-0.2cm}
\begin{align}
&\!\!\!\!\!\!\!\!\!\!\!\!s^+(\vpsi',\vpsi^+\!,\mPsi^-)\!=\!{\textstyle\frac{1}{\tau(\vpsi',\vpsi^+\!,\mPsi^-)}}\,\text{e}^{-\frac{||\vpsi'\!-\vpsi^+||_2^2}{2\sigma^2}} 
\!\text{ and }\;
s^-(\vpsi',\vpsi^+\!,\mPsi^-)\!=\!{\textstyle\frac{1}{\tau(\vpsi',\vpsi^+\!,\mPsi^-)}}\,\maxx\limits_{k=1,\cdots,K}\text{e}^{-\frac{||\vpsi'\!-\vpsi_k^-||_2^2}{2\sigma^2}}\!\!,\label{eq:pos_neg}
\end{align}

\vspace{-0.5cm}
\noindent where $\mPsi^-\!\!\equiv\!\{\vpsi_k^-\}_{k=1}^K$. Expression $\tau(\vpsi',\vpsi^+\!,\mPsi^-)$ given below normalizes probability partitions in Eq. \eqref{eq:pos_neg}: 
\vspace{-0.6cm}
\begin{align}
&\!\!\!\!\!\!\!\!\!\!\!\!\tau(\vpsi',\vpsi^+\!,\mPsi^-)\!=\!\text{e}^{-\frac{||\vpsi'\!-\vpsi^+||_2^2}{2\sigma^2}} \! +\! {\textstyle\sum}_{k=1}^K \,\text{e}^{-\frac{||\vpsi'\!-\vpsi_{k}^-||_2^2}{2\sigma^2}}\!,
\end{align}

\vspace{-0.3cm}
\noindent while $\sigma$ determines the steepness of likelihood partitions. The Soft Assignment step is performed by reweighting ${\vpsi'}_1^+,\cdots,{\vpsi'}_\eta^+$ by $s^+({\vpsi'}_1^+,\cdot,\cdot),\cdots,s^+({\vpsi'}_\eta^+,\cdot,\cdot)$ and ${\vpsi'}_{11}^-,\cdots,{\vpsi'}_{\eta K}^-$ by $s^-({\vpsi'}_{11}^-,\cdot,\cdot),\cdots,s^-({\vpsi'}_{\eta K}^-,\cdot,\cdot)$. Algorithm \ref{alg:neha} (with steps highlighted in blue) realizes NESA. NEHA and NESA use  combination losses: $L^\dagger\left(\vpsi^*;\vpsi^{+}\!,\mPsi^{-}\right)\!+\!\lambda L^\ddagger\left(\vmu,\mTheta;\;\vmu^{+}\!,\mTheta^{+}\!,\vmu^{-}\!,\mTheta^{-}\right)$.

\vspace{0.1cm}
\noindent{\textbf{NNO.}} Nearest Neighbor Only (NNO) strategy is given for completeness. 
NNO simply encourages the standard head with one FC layer (gray block in Figure \ref{fig:pipeline}) to get closer not only to target samples of MMD but also to the positive approximate nearest neighbor(s) retrieved from DeepFashion. NNO uses combined losses: $L^\dagger\left(\vpsi^*;\vpsi^{+}\!,\mPsi^{-}\right)\!+\!\lambda ||\vpsi^*\!-\frac{1}{\eta}\sum\nolimits_{n=1}^\eta{\vpsi'}_n^{+} ||_2^2$.


\vspace{0.1cm}
\noindent\textbf{\fontsize{11}{11}\selectfont Text-based Task.}\label{sec:tp} Figure \ref{fig:mm_decoder_t} shows that apart from the standard GRU decoder (gray blocks), we use translating network \cite{ott2018scaling} to translate \cite{ott2018scaling} ground truth sentences from English into French, German and Russian, with one GRU per language. For English, we have a GRU with hidden states $\vh^e_1,\cdots,\vh^e_{M'}$, output predictions $\vp^e_1,\cdots,\vp^e_{M'}$ and ground truth one-hot vectors $\vy^e_1,\cdots,\vy^e_{M'}$. By analogy, we use analogous streams for other languages. 
Moreover, every $\vp^e_m\!\in\!\mbr{7457}$ is an output of an FC layer connected to the corresponding hidden state $\vh^e_m\!\in\!\mbr{1024}$. The FC layer translates hidden states into word activation vectors corresponding to a 7457 dimensional dictionary. Note that for every language, the dictionary size differs. For French, we have 9519 words after considering words with the occurrence of at least $5\times$ given the training data. Each sentence starts with the start-of-sentence token, ends with the end-of-sentence token and is padded to the maximum sentence length of $M'\!=\!20$ with the pad-sentence token. Pink blocks realize the assisted supervision for the text-based task. At the test time, they are removed. The final loss for the Text Decoder becomes:
\vspace{-0.3cm}
\begin{align}
	&\!\!\!\!\!\!\!\!\!\!\!\!L\left(\left\{(\vp_m^e,\vy_m^e),\;(\vp_m^f,\vy_m^f),\cdots\right\}_{m=1}^{M'};\;\lambda^f,\cdots\right)\!=\!
	\sum\limits_{m=1}^{M'} {\vy_{m}^e}^T\text{log}\left(\vp^e_{m}\right) + \lambda^f {\vy_{m}^f}^T\text{log}\left(\vp^f_{m}\right)+\cdots\;,
	\end{align}

\vspace{-0.2cm}
\noindent where $\lambda^f$ is the relevance constant of the French translation task, and $\lambda^g$ and $\lambda^r$ are relevance constants for German/Russian but we omit them from notations for brevity.


\vspace{0.1cm}
\noindent\textbf{\fontsize{11}{11}\selectfont Task Discriminator (TD).}\label{sec:td} The Context Descriptor\footnote{For evaluations where we use TD, the Context Descriptor $\vpsi$ is in fact 2048 dimensional as its both halves are dedicated to text- and image-based tasks, respectively. For individual tasks, $\vpsi$ are 1024 dimensional.} 
$\vpsi\!\in\!\mbr{2048}$ is passed to an FC layer ($2048\!\times\!3$ size) following the cross-entropy loss with task labels: {\em text-based}, {\em image-based} and {\em text+image-based}. During training, we can access such labels. Thus, during testing, we can go beyond separate protocols of the baseline model \cite{saha2018towards}. Figure \ref{fig:dialogue} shows TD and the switches that pass relevant halves of $\vpsi$ to subsequent modules.

\begin{table}[t]
\centering
\parbox{0.51\textwidth}{
\renewcommand{\arraystretch}{0.5}
\setlength{\tabcolsep}{0.25em}
\fontsize{8}{8}\selectfont
\begin{tabular}{ccc c c}
\toprule
																																		&                                                  \kern-0.9em&\kern-0.9em             & BLEU           & NIST          \\ 
\midrule
\kern-0.9em\multirow{6}{*}{\rotatebox{90}{MMD {\em v1}}}\kern-0.3em & \kern-0.9emT-HRED 	\kern-0.9em&\kern-0.9em\cite{saha2018towards}       & 14.58          & 2.61          \\
																																		& \kern-0.9emM-HRED 																					\kern-0.9em&\kern-0.9em\cite{saha2018towards}       & 20.42          & 3.09          \\ 
																																		& \kern-0.9emM-HRED+attention 																\kern-0.9em&\kern-0.9em\cite{saha2018towards}       & 19.58          & 2.46          \\ 
																																		& \kern-0.9emM-HRED+attention+KB 														\kern-0.9em&\kern-0.9em\cite{agarwal2018knowledge}  & -              & -             \\
\cmidrule{2-5}
 & \kern-0.9em(Ours) Pre-training (French)\kern-0.9em & \kern-0.9em & 24.35 & 4.12\\
 & \kern-0.9em\textbf{(Ours) Assisted sup. (French)} \kern-0.9em&\kern-0.9em & \textbf{26.21} & \textbf{4.45} \\
\midrule
\kern-0.9em\multirow{8}{*}{\rotatebox{90}{MMD {\em v2}}}\kern-0.3em & \kern-0.9emT-HRED 	\kern-0.9em&\kern-0.9em\cite{saha2018towards} 			& 35.9           & 5.14          \\ 
																																		& \kern-0.9emM-HRED 																					\kern-0.9em&\kern-0.9em\cite{saha2018towards}       & 56.67          & 7.51          \\ 
																																		& \kern-0.9emM-HRED+attention 																\kern-0.9em&\kern-0.9em\cite{saha2018towards}       & 50.20          & 6.64          \\  
																																		& \kern-0.9emM-HRED+attention+KB 														\kern-0.9em&\kern-0.9em\cite{agarwal2018knowledge}  & 46.36          & -             \\
\cmidrule{2-5}
  & \revised{\kern-0.9em(Ours) Augmentation (random deletion)\kern-0.9em} & \kern-0.9em & \revised{56.83} & \revised{7.55}\\
  & \revised{\kern-0.9em(Ours) Augmentation (sentence compr. \cite{token_cnn})\kern-1.8em} & \kern-0.9em & \revised{57.65} & \revised{7.62}\\
  & \revised{\kern-0.9em(Ours) Augmentation (back translation  \cite{ott2018scaling})\kern-2em} & \kern-0.9em & \revised{59.06} & \revised{7.96}\\
  & \revised{\kern-0.9em(Ours) Pre-training (on SIMMC dataset \cite{crook2020simmc})\kern-2em} & \kern-0.9em & \revised{58.91} & \revised{7.95}\\
  & \revised{\kern-0.9em(Ours) Training on MMD+SIMMC)\kern-2em} & \kern-0.9em & \revised{59.03} & \revised{7.98}\\
& \kern-0.9em(Ours) Pre-training (French)   		          	\kern-0.9em&\kern-0.9em															& 58.78          & 7.91          \\ 
																																		& \kern-0.9em\textbf{(Ours) Assisted sup. (French)} 				\kern-0.9em&\kern-0.9em 														& \textbf{60.12} & \textbf{8.11} \\
																																		& \kern-0.9em{\fontsize{7}{7}\selectfont\textbf{(Ours) Assisted sup. (French+German)}} 		\kern-2.5em&\kern-0.9em 														& \textbf{60.51} & \textbf{8.17} \\
																																		& \kern-1.2em{\fontsize{7}{7}\selectfont\textbf{(Ours) Assisted sup. (French+German+Russian)}} 		\kern-2.2em&\kern-0.9em 														& \textbf{60.75} & \textbf{8.22} \\
\cmidrule{2-5}
\kern-0.9em\multirow{4}{*}{\rotatebox{90}{Tran.}}\kern-0.3em & \kern-0.9em(Ours) Pre-training (French)   		          	\kern-0.9em&\kern-0.9em			& 60.88          & 9.28          \\ 
																																		& \kern-0.9em\textbf{(Ours) Assisted sup. (French)} 				\kern-0.9em&\kern-0.9em 	& \textbf{64.47} & \textbf{11.18} \\
																																		& \kern-0.9em{\fontsize{7}{7}\selectfont\textbf{(Ours) Assisted sup. (French+German)}} \kern-2.5em&\kern-0.9em & \textbf{65.54} & \textbf{12.41} \\
																																		& \kern-1.2em{\fontsize{7}{7}\selectfont\textbf{(Ours) Assisted sup. (French+German+Russian)}} \kern-2.2em&\kern-0.9em & \textbf{66.19} & \textbf{12.89} \\
\bottomrule
\end{tabular}
\vspace{0.3cm}
\captionof{table}{Text-based task (MMD {\em v1} \& {\em v2}). T-HRED / M-HRED are text-only HRED / Multimodal HRED. {\em Tran.}: transformer backb. \cite{46201}.}
\label{tab:texttask}
\vspace{-0.3cm}
}
\hspace{0.2cm}
\parbox{0.46\textwidth}{
\centering
\renewcommand{\arraystretch}{0.5}
\setlength{\tabcolsep}{0.25em}
\fontsize{8}{8}\selectfont
\begin{tabular}{ccc c c c}
\toprule
												& \kern-0.9em                 								&                   										& R$@1$          & R$@2$          & R$@3$          \\
\midrule
\kern-0.9em\multirow{6}{*}{\rotatebox{90}{MMD {\em v1}}}\kern-0.3em & \kern-0.9emT-HRED\kern-0.9em								&\kern-0.9em\cite{saha2018towards}      & 46.0           & 64.0           & 75.0           \\  
																																		& \kern-0.9emM-HRED\kern-0.9em								&\kern-0.9em\cite{saha2018towards}      & 72.0           & 86.0           & 92.0           \\  
																																		& \kern-0.9emM-HRED+attention\kern-0.6em			&\kern-0.9em\cite{saha2018towards}      & 79.0           & 88.0           & 93.0           \\
\cmidrule{2-6}
																																		& \kern-0.9em(Ours) NNO $\eta\!=\!1$\kern-0.9em&         															& 82.6          & 88.8          & 93.2          \\
																																		& \kern-0.9em(Ours) NNO $\eta\!=\!2$*\kern-0.9em&         															& 83.0          & 88.9          & 93.2          \\
																																		& \kern-0.9em(Ours) NEHA $\eta\!=\!4$*				 &              													& 84.5          & 89.7          & 93.6          \\  
																																		& \kern-0.9em\textbf{(Ours) NESA $\eta\!=\!4$*}&							 													& \textbf{85.3} & \textbf{90.3} & \textbf{94.0} \\
\midrule
\kern-0.9em\multirow{6}{*}{\rotatebox{90}{MMD {\em v2}}}\kern-0.3em & \kern-0.9emT-HRED 	\kern-0.9em							&\kern-0.9em\cite{saha2018towards}      & 44.0           & 60 .0          & 72.0         \\  
																																		& \kern-0.9emM-HRED \kern-0.9em								&\kern-0.9em\cite{saha2018towards}      & 69.0           & 85.0           & 90.0          \\  
																																		& \kern-0.9emM-HRED+attention\kern-0.6em			&\kern-0.9em\cite{saha2018towards}      & 78.0           & 87.0          & 92.3          \\
\cmidrule{2-6}
																																		& \kern-0.9em(Ours) NNO $\eta\!=\!1$\kern-0.9em&           														& 82.5          & 88.6          & 92.8          \\
																																		& \kern-0.9em(Ours) NNO $\eta\!=\!2$*\kern-0.9em&           														& 83.1          & 88.8          & 92.9          \\
																																		& \kern-0.9em(Ours) NEHA $\eta\!=\!4$* 			  &              													& 84.5          & 89.5          & 93.2          \\  
																																		& \kern-0.9em\textbf{(Ours) NESA $\eta\!=\!4$*}&																				& \textbf{85.2} & \textbf{90.1} & \textbf{93.7} \\
\bottomrule
\end{tabular}
\vspace{0.3cm}
\captionof{table}{Image-based task (MMD {\em v1} \& {\em v2}) for one positive and $K\!=\!5$ negative target images. T-HRED is HRED with context images ignored in training. M-HRED is the Multimodal HRED. See Recall at top-$1$, $2$ and $3$, `*' is the optimal $\eta$.}
\label{tab:imtask}
\vspace{-0.4cm}
}
\end{table}

\vspace{-0.1cm}
\section{Experiments}
\label{sec:ex}



\noindent\textbf{\fontsize{11}{11}\selectfont Datasets}. Our experiments are conducted on the MMD datasets \cite{saha2018towards} {\em v1} and {\em v2} containing $\sim$150000 dialogues and the SIMMC dataset \cite{crook2020simmc}, with $\sim$13K human-human dialogues and $\sim$169K utterances. The assisted supervision for the text-based task is achieved via model \cite{ott2018scaling} trained on the WMT \cite{bojar-EtAl:2014:W14-33} and Paracrawl \cite{aeb1138d856e477a9ea0f3ee5900cab1} datasets containing $\sim$150M sentence pairs. The assisted supervision for the image-based task is achieved by retrieving relevant feature descriptors from the  DeepFashion dataset \cite{liu2016deepfashion} ($\sim$0.8M images).

\vspace{0.1cm}
\noindent{\textbf{MMD}} dataset \cite{saha2018towards} contains 105439 train, 22595 validation and 22595 test dialogues, each with $\sim$40 shopper-retailer utterances containing a sentence, images or both  modalities. 
We used train, validation and test splits to train, select hyperparameters and report final results, respectively.  MMD {\em v2} does not contain additional image descriptions  from the agent.

\vspace{0.1cm}
\noindent{\textbf{SIMMC}} dataset \cite{crook2020simmc} has  $\sim$13K human-human dialogs and $\sim$169K utterances, it uses  a multimodal Wizard-of-Oz (WoZ) setup, on two shopping domains, furniture (grounded in a shared virtual environment) and fashion (grounded in an evolving set of images).

\vspace{0.1cm}
\noindent\textbf{\fontsize{11}{11}\selectfont Settings.} 
Following Saha \etal \cite{saha2018towards}, we perform the text- and image-based tasks for which we use the same hidden unit size, text encoding size and the learning rate as M-HRED \cite{saha2018towards}. For our combined task (TD module), the hidden unit size is doubled (Section \ref{sec:td}). For the text- and image-based tasks, we report  BLEU/NIST 
\cite{saha2018towards} and Recall at top-$l$ cut-off (R$@l$).


\vspace{0.1cm}
\noindent\textbf{\fontsize{11}{11}\selectfont Results}. 
Below we start with cross-validation of key hyperparameters followed by presenting our main results for text-, image- and mixed (text+images) tasks. 

\vspace{0.1cm}
\noindent{\textbf{Cross-validation of $\lambda^f$ and $\lambda$.}} 
\revised{For joint training of French auxiliary decoder with the base English decoder,  we cross-validated $\lambda^f\!\in\!\{0.1, 0.3, 0.5, 0.7, 1.0\}$ on the validation set (see Figure \ref{fig:cv1}). We fixed $\lambda^f\!\!=\!0.3$ throughout experiments as this value yielded the highest score of \textbf{60.25\%} (BLEU) on the MMD {\em v2} validation split. If we use two auxiliary decoders \eg, French and German, we set $\lambda^f\!=\!\lambda^g\!=\!0.15$. For three auxiliary decoders, we set $\lambda^f\!=\!\lambda^g\!=\!\lambda^r\!=\!0.1$.} 
For joint training of the main stream (FC$\rightarrow$ReLU$\rightarrow$FC) and the assisted supervision stream in Feature Matching Head from Figure \ref{fig:pipeline}, we set $\lambda\!=\!0.5$ following cross-validation on the validation set given NEHA, shown Figure \ref{fig:cv2}.

\comment{
\begin{table}[t]
\vspace{-0.2cm}
\centering
\begin{tabular}{cc|c||c|c|}
\kern-0.9em&\kern-0.9em             & BLEU  & & R$@1$                 \\
\hline
\kern-0.9em{\fontsize{7}{7}\selectfont M-HRED} \kern-0.9em&\kern-0.9em 	& 52.17 & \kern-0.3em{\fontsize{7}{7}\selectfont M-HRED+att.}														\kern-0.6em& 75.05\\\hline
\kern-0.9em{ Assisted sup. (French)} 										\kern-0.9em&\kern-0.9em 	& 55.29  & \kern-0.3em NEHA\kern-0.4em & 81.50  \\  
\kern-0.6em{\fontsize{7}{7}\selectfont Assisted sup. (French+German+Russian)} 		\kern-0.9em&\kern-0.9em 	& \textbf{56.11}  & \kern-0.3em NESA\kern-0.4em	& \textbf{82.43}  \\
\hline
\end{tabular}
\caption{Mixed  task on the MMD dataset ({\em v2}). We use the assisted supervision for both text- and image-based tasks, 
with the Task Discriminator.}
\label{tab:mixed}
\vspace{-0.1cm}
\end{table}
}

\begin{table}[t]
\centering
\parbox{0.49\textwidth}{
\vspace{-0.6cm}
\renewcommand{\arraystretch}{0.5}
\setlength{\tabcolsep}{0.25em}
\fontsize{8}{8}\selectfont
\centering
\begin{tabular}{cc c  c c}
\toprule
\kern-0.9em&\kern-0.9em             & BLEU  & & R$@1$                 \\
\midrule
\kern-0.9em{\fontsize{7}{7}\selectfont M-HRED} \kern-0.9em&\kern-0.9em 	& 52.17 & \kern-0.1em{\fontsize{7}{7}\selectfont M-HRED+att.}														\kern-0.6em& 75.05\\
\midrule
\kern-0.9em{ Assisted sup. (French)} 										\kern-0.9em&\kern-0.9em 	& 55.29  & \kern-0.3em NEHA\kern-0.4em & 81.50  \\  
\kern-0.6em{\fontsize{7}{7}\selectfont (French+German+Russian)} 		\kern-0.2em&\kern-0.9em 	& \textbf{56.11}  & \kern-0.3em NESA\kern-0.4em	& \textbf{82.43}  \\
\bottomrule
\end{tabular}
\vspace{0.3cm}
\caption{Mixed  task (MMD {\em v2}) with the Task Discr., assisted supervision (text and images). 
}
\label{tab:mixed}
\vspace{0.1cm}
\begin{tabular}{c ccc c}
\toprule
\kern-0.9em& User 1  &  User 2   & User 3  & mean             \\
\midrule
clarity&  61.6   &  58.4   & 64.2  & 61.4            \\
compactness&  52.0 & 52.8 &  54.6   &   53.1             \\
helpfulness&  62.0   &  60.2   &  63.0 & 61.7             \\
\bottomrule
\end{tabular}
\vspace{0.3cm}
\caption{User study on the mixed  task (MMD dataset {\em v2}). Our approach \vs M-HRED.}
\label{tab:user}
\vspace{0.1cm}
{\renewcommand{\arraystretch}{0.5}
\setlength{\tabcolsep}{0.25em}
\fontsize{8}{8}\selectfont
\centering
\begin{tabular}{cc}
\toprule
Component (or method)  &  runtime (h)             \\
\midrule
T-HRED / M-HRED &   15 / 15            \\
Pre-training (Fr) (+fine-tuning En) & 16 + 6 \\
Augmentation (back translation) + transl.\kern-1.1em & 15 + 40\\
Assisted sup. (Fr / Fr+Ge / Fr+Ge+Ru) & 20 / 29 / 38\\
\midrule
Translator \cite{ott2018scaling} (En $\rightarrow$ Fr / Ge / Ru) & 20 / 20 / 20\\
Evaluating BLEU \& NIST & 0.5\\
\bottomrule
\end{tabular}
}
\vspace{0.3cm}
\caption{\revised{Runtimes: text task (MMD {\em v2}).}}
\label{tab:time1}
\vspace{-0.7cm}
}
\hspace{0.3cm}
\parbox{0.47\textwidth}{
\renewcommand{\arraystretch}{0.5}
\setlength{\tabcolsep}{0.25em}
\fontsize{8}{8}\selectfont
\centering
\begin{tabular}{l llll}
\toprule
                                         & BLEU  & R@1  & R@5  & {R@10} \\
\midrule
HRE (SIMMC)		 & 0.079 & 16.3 & 33.1 & {41.7} \\
\midrule
Ours F+R+G                         & \textbf{0.102} & n/a    & n/a    & {n/a}    \\
Ours+Trans. F+R+G                         & \textbf{0.187} & n/a    & n/a    & {n/a}    \\
\cmidrule[0.5pt](rl){1-5}
Ours NEHA                              & n/a     & 17.3 & 33.7 & {42.2} \\
Ours NESA                              & n/a     & \textbf{20.1} & \textbf{35.5} & {\textbf{43.1}} \\ 
\bottomrule
\end{tabular}
\vspace{0.3cm}
\caption{SIMMC-Fashion (Task 2). Response Generation. {\em F+R+G} are French, Russian and German auxiliary tasks. {\em Tran.} is the transformer backbone \cite{46201}.}
\label{sec:simmc}
\vspace{0.1cm}
{
\renewcommand{\arraystretch}{0.5}
\setlength{\tabcolsep}{0.25em}
\fontsize{8}{8}\selectfont
\centering
\begin{tabular}{cc}
\toprule
Component (or method)  &  runtime (h)             \\
\midrule
T-HRED / M-HRED &   15 / 15\\
NNO & 16\\
NEHA / NESA & \kern-0.9em 18 / 19\\
\midrule
ResNet-50 features (MMD+DeepFashion)\kern-1.1em &   6            \\
FAISS search \cite{johnson2019billion} (+SVD) &  1.5 (+2) \\
Evaluating R@1 &  0.1 \\
\midrule
T-HRED / M-HRED (text+image) & 30 / 30 \\
Mixed task (text+image+task discr.) & 40 \\
\bottomrule
\end{tabular}
\vspace{0.3cm}
\caption{\revised{Runtimes: image-based task (various comp.) and the mixed task (MMD {\em v2}).}}
\label{tab:time2}
\vspace{-0.2cm}
}
}
\end{table}

\comment{
\begin{table}[t]
\centering
\begin{tabular}{l|l|l|l|l|l|l|l|l|l|l|l|}
                                         & BLEU  & R@1  & R@5  & \multicolumn{8}{l|}{R@10} \\ \hline
HRE - (SIMMC)		 & 0.079 & 16.3 & 33.1 & \multicolumn{8}{l|}{41.7} \\ \hline
Ours - F+R+G                         & \textbf{0.102} & n/a    & n/a    & \multicolumn{8}{l|}{n/a}    \\
Ours+Trans. - F+R+G                         & \textbf{0.187} & n/a    & n/a    & \multicolumn{8}{l|}{n/a}    \\
\hline\hline
Ours - NEHA                              & n/a     & 17.3 & 33.7 & \multicolumn{8}{l|}{42.2} \\
Ours - NESA                              & n/a     & \textbf{20.1} & \textbf{35.5} & \multicolumn{8}{l|}{\textbf{43.1}} \\ \hline

\end{tabular}
\caption{SIMMC-Fashion (Task 2). Response Generation.}
\label{sec:simmc}
\vspace{-0.3cm}
\end{table}
}

\vspace{0.1cm}
\noindent{\textbf{Image-based Task.}}  Firstly,  we evaluate the baseline M-HRED+attention with ResNet-50 in place of VGG-16, and we note that the results are within $\pm$0.3\% of results given the original M-HRED+attention with VGG-16. Table \ref{tab:imtask} shows that using the assisted supervision via the NNO strategy with one nearest neighbor ($\eta\!=\!1$) improves results over the baseline M-HRED+attention by $\sim$3.6\% and $\sim$4.5\% (R$@1$) given versions {\em v1} and {\em v2} of the MMD dataset. Choosing the optimal number of nearest neighbors for NNO ($\eta\!=\!2$) improves results by further 0.4\% (R$@1$) over NNO ($\eta\!=\!1$) on both versions of MMD. Moreover, utilizing our subspace-based NEHA, we obtain 5.5\% and 5.5\% (R$@1$) improvement over  the baseline M-HRED+attention given  both versions of MMD. Our best performer, subspace-based NESA yields \textbf{6.3\%} and \textbf{7.2\%} (R$@1$) improvement over the baseline M-HRED+attention model. 

\comment{
\begin{table}[t]
\centering
\parbox{0.48\textwidth}{
\renewcommand{\arraystretch}{0.5}
\setlength{\tabcolsep}{0.25em}
\fontsize{8}{8}\selectfont
\centering
\begin{tabular}{cc}
\toprule
Component (or method)  &  runtime (h)             \\
\midrule
T-HRED / M-HRED &   15 / 15            \\
Pre-training (Fr) (+fine-tuning En) & 16 + 6 \\
Augmentation (back translation) + transl.\kern-1.1em & 15 + 40\\
Assisted sup. (Fr / Fr+Ge / Fr+Ge+Ru) & 20 / 29 / 38\\
\midrule
Translator \cite{ott2018scaling} (En $\rightarrow$ Fr / Ge / Ru) & 20 / 20 / 20\\
Evaluating BLEU \& NIST & 0.5\\
\bottomrule
\end{tabular}
\vspace{0.3cm}
\caption{\revised{Runtimes on text-based task (various components on the MMD dataset {\em v1}).}}
\label{tab:time1}
\vspace{-0.4cm}
}
\hspace{0.1cm}
\parbox{0.48\textwidth}{
\renewcommand{\arraystretch}{0.5}
\setlength{\tabcolsep}{0.25em}
\fontsize{8}{8}\selectfont
\centering
\begin{tabular}{cc}
\toprule
Component (or method)  &  runtime (h)             \\
\midrule
T-HRED / M-HRED &   15 / 15\\
NNO & 16\\
NEHA / NESA & \kern-0.9em 18 / 19\\
\midrule
ResNet-50 features (MMD+DeepFashion)\kern-1.1em &   6            \\
FAISS search \cite{johnson2019billion} (+SVD) &  1.5 (+2) \\
Evaluating R@1 &  0.1 \\
\midrule
T-HRED / M-HRED (text+image) & 30 / 30 \\
Mixed task (text+image+task discr.) & 40 \\
\bottomrule
\end{tabular}
\vspace{0.3cm}
\caption{\revised{Runtimes: image-based task (various comp. on MMD {\em v1}) and the Mixed task.}}
\label{tab:time2}
\vspace{-0.4cm}
}
\end{table}
}

\vspace{0.1cm}
\noindent{\textbf{Text-based Task.}} 
Table \ref{tab:texttask} shows results (BLEU and NIST metrics) by comparing target sentences against predicted sentences. Pre-training Text Decoder with French language prior to fine-tuning on English improves results by $\sim$4\% and $\sim$2.1\% (BLEU) over the M-HRED baseline on both MMD {\em v1} and {\em v2}. \revised{Using random word deletions for augmentation yielded gain of 0.16\% (BLEU) over the M-HRED baseline (MMD {\em v2}). Augmentations via so-called sentence compression \cite{token_cnn} scored $\sim$1\% over M-HRED, whereas augmentations via the so-called back-translation (using translating model \cite{ott2018scaling}) scored $\sim$2.4\% over M-HRED. Pre-training on SIMMC \cite{crook2020simmc}  was marginally worse (and very similar to combined training on MMD+SIMMC). However, using the assisted supervision, that is, an auxiliary decoder for French, improves results by further $\sim$3.5\% (BLEU) over the M-HRED baseline (MMD {\em v2}). Augmentations by back translation require translating sentences twice English$\rightarrow$French$\rightarrow$English (additional 20 hours), whereas our assisted supervision requires only English$\rightarrow$French translation.}  Adding auxiliary German and Russian decoders (to French) and the main decoder for English yields over $\mathbf{4}\%$ (BLEU) over the M-HRED baseline (MMD {\em v2}). 
Finally, using the transformer backbone \cite{46201} results in a $\sim$5\% boost. The benefit of adding multiple auxiliary language decoders is clear. In what follows, we use the GRU backbone not transformers 
(the backbone choice is a secondary matter). 
\revised{Pre-training the text backbone on the SIMMC dataset \cite{crook2020simmc} before applying our assisted step may also boost results. Applying the sentence compression model \cite{token_cnn} via an auxiliary decoder (in addition to French, German and Russian) in our assisted supervision is also possible.}

\vspace{0.1cm}
\noindent{\textbf{Mixed Task.}} Firstly, we evaluate our Task Discriminator on the  MMD dataset ({\em v2}) and note that it achieves 97.0\% accuracy. This means that results in Tables \ref{tab:texttask} and \ref{tab:imtask} represent upper bound scores for this paragraph as both tables report on two separate tasks (oracle knowledge regarding which task is which) according to protocol in Saha \etal \cite{saha2018towards}. Table \ref{tab:mixed} shows that results for the mixed task dropped marginally compared to results in Tables \ref{tab:texttask} and \ref{tab:imtask}. Our best assisted supervision methods outperformed best baselines M-HRED and M-HRED+attention equipped with Task Discriminator by $\sim$4\% (BLEU) and $\sim$7.4\% (R$@1$).

\comment{
\begin{figure}[t]
\centering
%
\begin{subfigure}[t]{0.495\linewidth}
\centering\includegraphics[width=4.5cm,trim=20 190 -10 320]{figures/cv3.pdf}
\caption{\label{fig:cv3}}
\end{subfigure}
\begin{subfigure}[t]{0.495\linewidth}
\centering\includegraphics[width=4.5cm,trim=20 190 -10 320]{figures/cv4.pdf}
\caption{\label{fig:cv4}}
\end{subfigure}
%
%
\caption{Fig. \ref{fig:cv3} \& \ref{fig:cv4} show cross-validation results (R$@1$ \& R$@3$ metric) on the validation set \wrt $\eta'$ and $\eta$ for the image-based task. If $\eta'\!=\!0$, we use $\mu^-$ and $\mu^+$ only.
}
\label{fig:cv34}
\vspace{-0.2cm}
\end{figure}
}

\vspace{0.1cm}
\noindent{\textbf{Ablations on NEHA \wrt $\eta'\!\!\leq\!\eta$.}} Below we investigate the impact of subspace size \wrt $\eta'$ and the impact of $\eta$ nearest neighbors retrieved from DeepFashion on the performance of image-based task.  Figure \ref{fig:cv3} shows that the best performance is attained for $\eta'\!=\!\eta\!=\!4$ and the trend suggests that $\eta'\!\approx\!\eta$ is a good choice. Figure \ref{fig:cv4} shows that $\eta'\!=\!\eta\!=\!5$ is a better choice for R$@3$, which allows two incorrect matches precede the correct one. Thus, including more nearest neighbors of positive/negative target images of MMD boosts the score.

\vspace{0.1cm}
\noindent{\textbf{Nearest Neighbors+the Hinge Loss.}} Positive/negative nearest neighbors retrieved from DeepFashion for positive/negative target images can be fed directly into our assisted supervision loss in Eq. \eqref{eq:assist}. Figures \ref{fig:cv3} and \ref{fig:cv4} evaluate such a setting ($\eta'\!\!=\!0$) as it is a special case of our subspace-based approach if $\eta'\!\!=\!0$ (only $\mu^-$ and $\mu^+$ are used if $\eta'\!\!=\!0$). On average, such a setting is $\sim$2\% worse than the subspace-based context. Subspaces capture robustly second-order statistics by discarding eigenvalue scaling and the smallest factors. 

\vspace{0.1cm}
\noindent{\textbf{User study.}} We asked 3 users to score our best performer \vs M-HRED on MMD (v2) (randomized test) in terms of {\em clarity}, {\em compactness} and {\em helpfulness} on 500 system responses. Table  \ref{tab:user} shows that  $\sim$61.0\% responses of the assisted supervision were clearer and more helpful (\vs 39\% of M-HRED). Both methods were generating similarly compact responses.

\vspace{0.1cm}
\noindent{\textbf{SIMMC.}} Table \ref{sec:simmc}  shows that using the multilingual decoding head yields 2.3\% and $\sim$10\% gain (BLEU) on RNN and transformers backbone over the HRE baseline (see the SIMMC paper for details of HRE). Moreover, our visual NESA yielded $\sim$4\% gain (R@1 score).

\revised{
\vspace{0.1cm}
\noindent{\textbf{Runtimes.}} 
Our code is implemented in PyTorch and evaluated on an NVIDIA Tesla P100 (unless stated otherwise). 
Table \ref{tab:time1} (runtimes for the text-based tasks) shows that the T-HRED and M-HRED baselines take $\sim$15 hours to train. Our assisted supervision (French) uses extra 5 hours. 
Translations are obtained off-line with translator \cite{ott2018scaling}. However, the best augmentation strategy that we tried (back translation) takes 55 hours, whereas our assisted superv. takes 40 hours (including translation time). Table \ref{tab:time2} (runtimes for the image-based tasks) shows that the T-HRED and M-HRED baselines take $\sim$15 hours to train. Nearest Neighbor Only, NEHA and NESA require 1, 3, and 4 extra hours. The off-line pre-processing includes the ResNet-50 feature extraction from MMD and DeepFashion (6 hours), nearest neighbor search with FAISS \cite{johnson2019billion} (1.5 hours, 4 GPUs) and running SVD (2 hours, 4 GPUs).}

\vspace{-0.4cm}
\section{Conclusions}
\label{sec:con}

We have introduced the assisted supervision which boosts the performance by leveraging AUDITED. Sampling auxiliary nearest neighbors from the natural manifold of fashion images helps create a meaningful visual context for the image task. With appropriate Soft Assignment reweighting and subspace modeling, benefits become clear while (by design) not posing any extra burden at the testing time. Learning to decode target dialogue sentences in several languages helps reduce the noise of single language syntactic. 

\vspace{0.1cm}
\noindent{\textbf{Acknowledgement.}} This work was supported by CSIRO’s Machine Learning and Artificial Intelligence Future Science Platform (MLAI FSP).

\appendix 
\addcontentsline{toc}{chapter}{APPENDICES}

\definecolor{beaublue}{rgb}{0.93, 0.97, 0.97}
\definecolor{blackish}{rgb}{0.2, 0.2, 0.2}

\definecolor{beaublue2}{rgb}{0.84, 0.9, 0.95}
\definecolor{blackish2}{rgb}{0.2, 0.2, 0.2}

\newcommand{\CO}{\color{black!40!blue}}
\newcommand{\COO}{\color{black!10!blue}}
\newcommand{\ROO}{\color{black!10!red}}

\section{Examples of Dialogues}

Below, we present three randomly selected dialogues. Responses generated by our system, denoted by ({\em our\_pred}), appear to make fewer mistakes and be more concise.

\vspace{0.2cm}
\noindent{\textbf{Example I:}}

\noindent{\fontsize{9}{9}\selectfont
User: {\em Show me something like the one in 4th image.}\\
Agent: {\em  The similar looking ones are img\_201, img\_320, img\_513.}\\
User: {\em show me something that will pair well with the 2nd image?}\\
Target\_A: {\em It can go well with black colors sports shoes.}\\
{\CO MMD\_pred:} {\em It can go well with regular style, regular fit, solid and regular style, solid and regular.}\\
{\CO Our\_pred:} {\em It can go well with {\COO casual fitted footwear.}}
}

\vspace{0.2cm}
\noindent{\textbf{Example II:}}

\noindent{\fontsize{9}{9}\selectfont
    User:  {\em and what about for the 1st image?}\\
    Agent:  {\em Image from the front, back and left directions respectively}\\
    User: {\em  Which of the celebrities usually wear watch similar looking to the one in the 2nd image?}\\
    Target\_A: {\em Celebrities cel\_278,cel\_255 and cel\_444 endorse this kind of {\COO watch}.}\\
    {\CO MMD\_pred:} {\em Celebrities cel\_30 and cel\_252 endorse this type of {\ROO handbag}.}\\
    {\CO Our\_pred:} {\em Celebrities cel\_60 and cel\_255 endorse this kind of {\COO watch}.}
}

\vspace{0.2cm}
\noindent{\textbf{Example III:}}

\noindent{\fontsize{9}{9}\selectfont
    Agent: {\em May I please ask your gender.}\\
    User: {\em Female, I am a 15 yr old female.}\\
    Agent: {\em Ok, this is helpful.}\\
    Target\_A: {\em I'll just take a few seconds to go through my catalog of items.}\\
    {\CO MMD\_pred:} {\em Absolutely. Please wait a bit while I search through my catalog quickly.}\\
    {\CO Our\_pred:} {\em Good! Just wait a few seconds while I go through my item catalog.}
}    

%
%

\section{Cross-validation of NESA}

\begin{figure}[!ht]
\vspace{0.5cm}
\centering
\includegraphics[height=5.0cm,trim=20 190 -10 260]{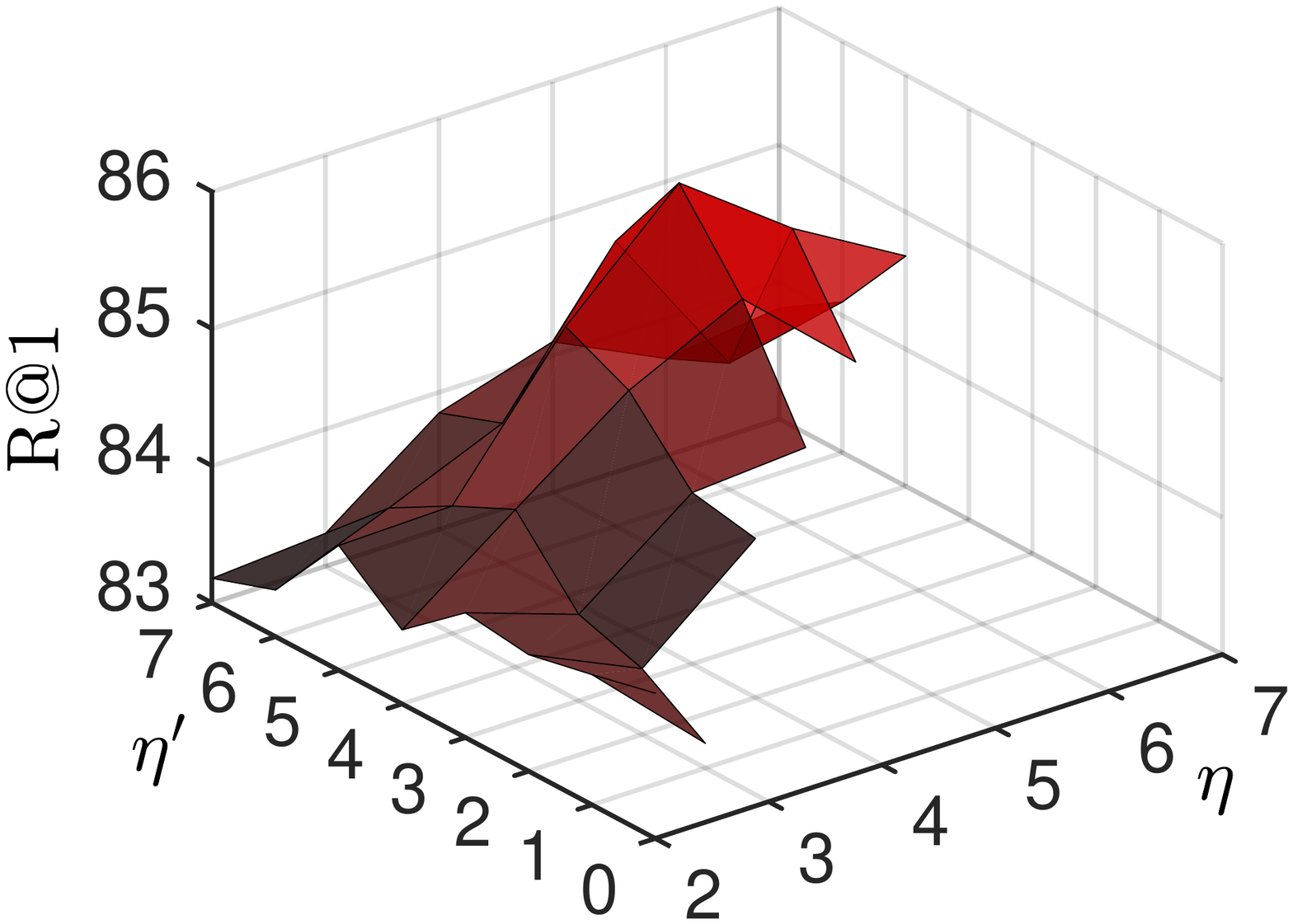}
\vspace{0.2cm}
\caption{Performance (R@1) w.r.t. the number of nearest neighbors $\eta$ and the number of leading eigenvectors $\eta'$ on NESA ($\sigma\!=\!0.5$).}
\label{fig:nesa_nnei}
\end{figure}

\begin{figure}[!ht]
\centering
\includegraphics[height=5.0cm,trim=20 190 -10 260]{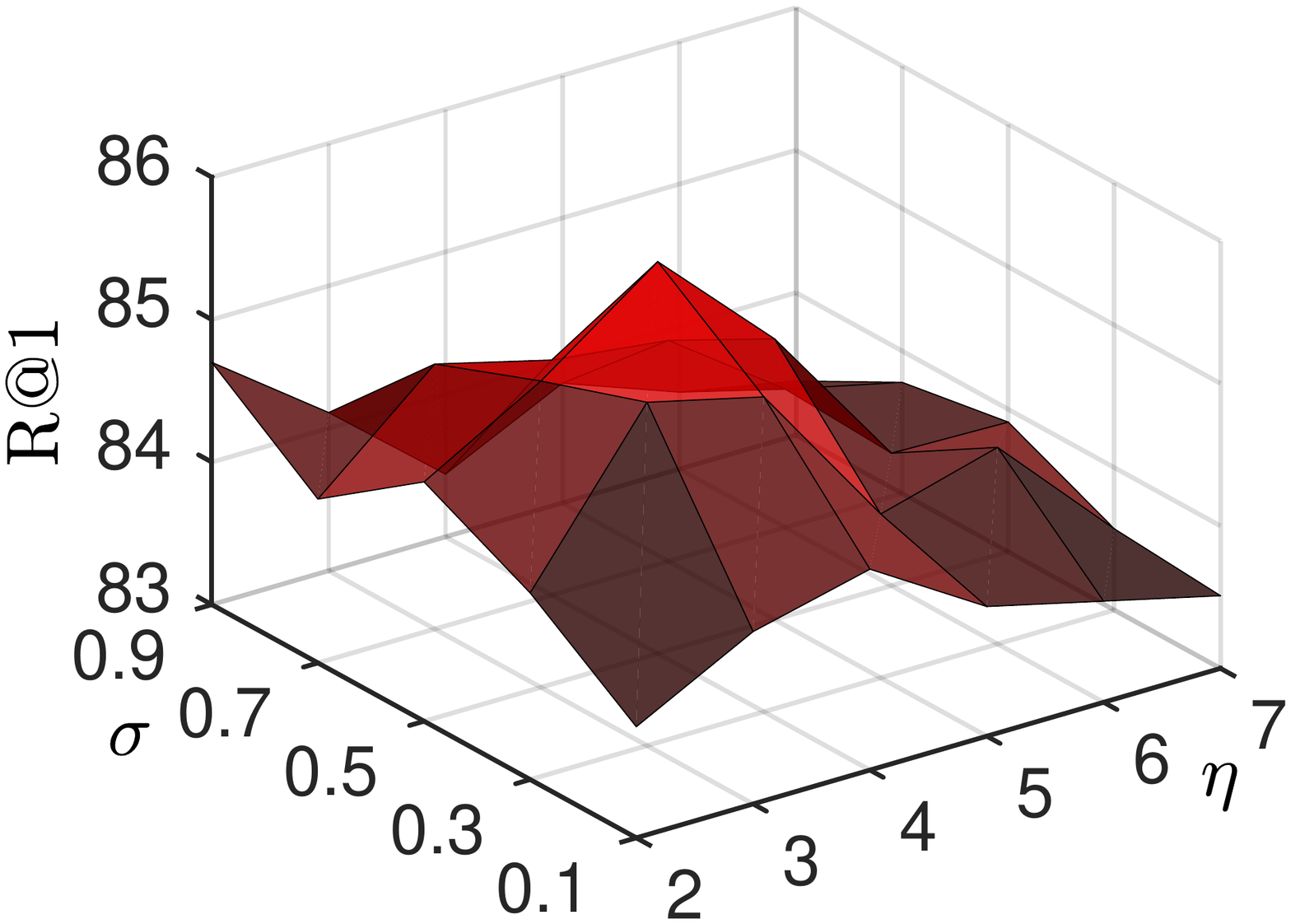}
\vspace{0.2cm}
\caption{Performance (R@1) w.r.t. the number of  nearest neighbors $\eta$ and the RBF bandwidth $\sigma$ on NESA ($\eta'\!=\!2$).}
\label{fig:nesa_nnsg}

\end{figure}

\begin{figure}[!ht]
\centering
\includegraphics[height=5.0cm,trim=20 190 -10 260]{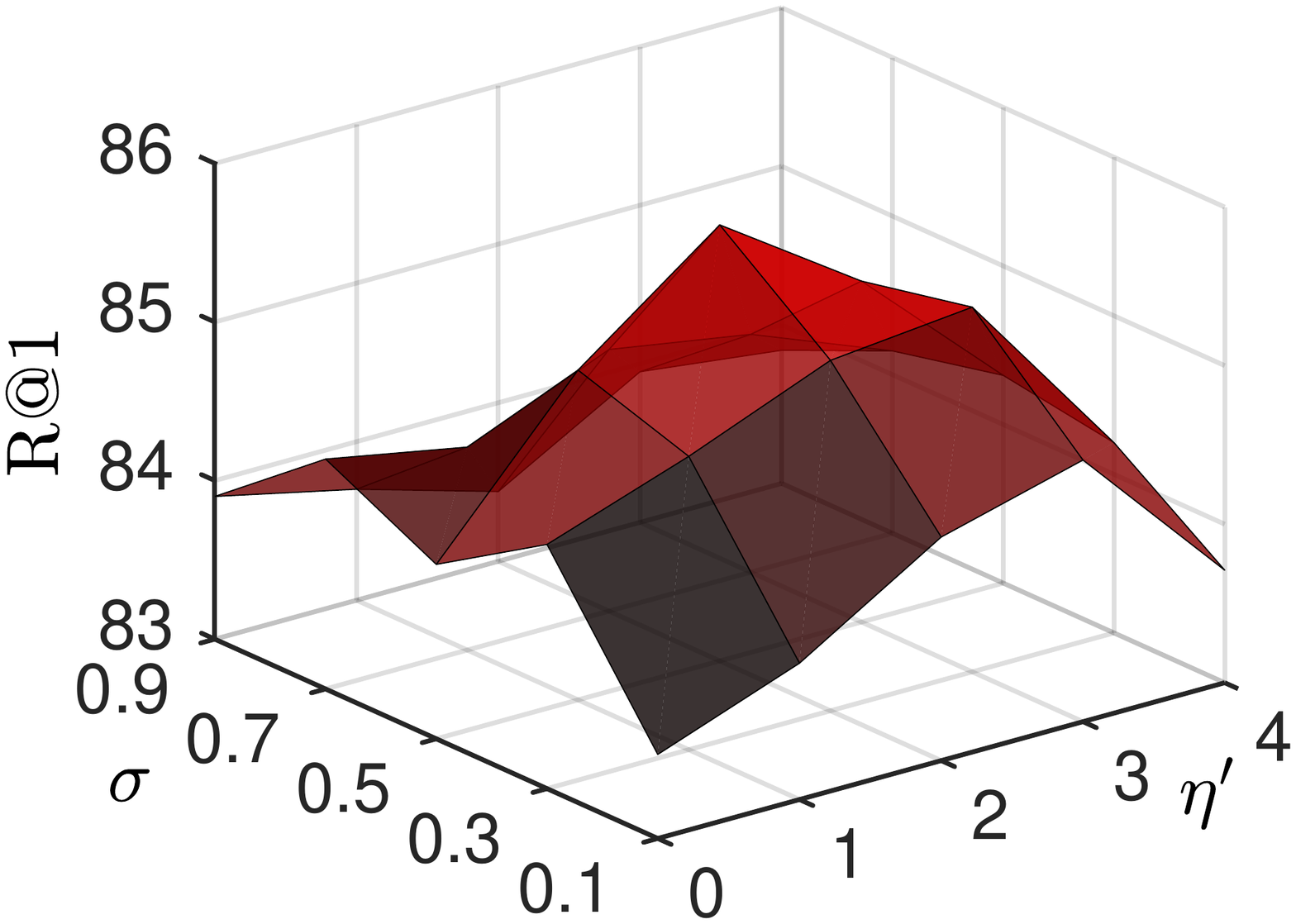}
\vspace{0.2cm}
\caption{Performance (R@1) w.r.t. the number of leading eigenvectors $\eta'$ and the RBF bandwidth $\sigma$ on NESA ($\eta\!=\!4$).}
\label{fig:nesa_eisg}
\end{figure}

Figures \ref{fig:nesa_nnei}, \ref{fig:nesa_nnsg} \ref{fig:nesa_eisg} evaluate the performance of  NESA model with respect to its parameters given the validation split. 

In Figure \ref{fig:nesa_nnei}, the bandwidth of the RBF kernel $\sigma$ value is fixed to 0.5. We vary the number of nearest neighbors $\eta\in\{2, 3, 4, 5, 6, 7\}$ and the number of leading eigenvectors $\eta'\in\{0, 1, 2, 3, 4, 5, 6, 7\}$.

In Figure \ref{fig:nesa_nnsg}, number of eigenvectors is fixed to 2, that is $\eta'=2$. We cross-validate the number of nearest neighbors $\eta\in\{2, 3, 4, 5, 6, 7\}$ and the RBF bandwidth set to $\sigma\in\{0.1, 0.3, 0.5, 0.7, 0.9\}$. 

In Figure \ref{fig:nesa_eisg}, number of nearest neighbors $\eta$ is fixed to 4. We cross-validate the number of leading eigenvectors $\eta'\in\{0, 1, 2, 3, 4\}$ and the RBF bandwidth $\sigma\in\{0.1, 0.3, 0.5, 0.7, 0.9\}$.

As previously indicated, the best $\eta'\approx\eta$. The best $\sigma=0.5$ which we use across all our experiments. Moreover, as $\eta'\approx\eta=4$, this indicates that our NESA can create a rich visual context for target images (much better context than directly forcing  target images to be close to their nearest neighbors in DeepFashion). Moreover, NESA outperforms NEHA (Figure 8, main submission).


{
\small
\bibliography{arxiv_version}

\begin{thebibliography}{63}
\providecommand{\natexlab}[1]{#1}
\providecommand{\url}[1]{\texttt{#1}}
\expandafter\ifx\csname urlstyle\endcsname\relax
  \providecommand{\doi}[1]{doi: #1}\else
  \providecommand{\doi}{doi: \begingroup \urlstyle{rm}\Url}\fi

\bibitem[Agarwal et~al.(2018)Agarwal, Dusek, Konstas, and
  Rieser]{agarwal2018knowledge}
Shubham Agarwal, Ondrej Dusek, Ioannis Konstas, and Verena Rieser.
\newblock A knowledge-grounded multimodal search-based conversational agent.
\newblock \emph{arXiv preprint arXiv:1810.11954}, 2018.

\bibitem[Agrawal et~al.(2015)Agrawal, Carreira, and Malik]{agrawal2015learning}
Pulkit Agrawal, Joao Carreira, and Jitendra Malik.
\newblock Learning to see by moving.
\newblock In \emph{Proceedings of the IEEE international conference on computer
  vision}, pages 37--45, 2015.

\bibitem[Ameixa et~al.(2014)Ameixa, Coheur, Fialho, and
  Quaresma]{ameixa2014luke}
David Ameixa, Luisa Coheur, Pedro Fialho, and Paulo Quaresma.
\newblock Luke, i am your father: dealing with out-of-domain requests by using
  movies subtitles.
\newblock In \emph{International Conference on Intelligent Virtual Agents},
  pages 13--21. Springer, 2014.

\bibitem[Antol et~al.(2015)Antol, Agrawal, Lu, Mitchell, Batra,
  Lawrence~Zitnick, and Parikh]{antol2015vqa}
Stanislaw Antol, Aishwarya Agrawal, Jiasen Lu, Margaret Mitchell, Dhruv Batra,
  C~Lawrence~Zitnick, and Devi Parikh.
\newblock Vqa: Visual question answering.
\newblock In \emph{Proceedings of the IEEE international conference on computer
  vision}, pages 2425--2433, 2015.

\bibitem[Banchs(2012)]{banchs2012movie}
Rafael~E Banchs.
\newblock Movie-dic: a movie dialogue corpus for research and development.
\newblock In \emph{Proceedings of the 50th Annual Meeting of the Association
  for Computational Linguistics (Volume 2: Short Papers)}, pages 203--207,
  2012.

\bibitem[Ba{\~n}{\'o}n et~al.(2020)Ba{\~n}{\'o}n, Chen, Haddow, Heafield,
  Hoang, Espl{\`a}-Gomis, Forcada, Kamran, Kirefu, Koehn, Ortiz-Rojas, Pla,
  Ram{\'i}rez-S{\'a}nchez, Sarr{\'i}as, Strelec, Thompson, Waites, Wiggins, and
  Zaragoza]{aeb1138d856e477a9ea0f3ee5900cab1}
Marta Ba{\~n}{\'o}n, Pinzhen Chen, Barry Haddow, Kenneth Heafield, Hieu Hoang,
  Miquel Espl{\`a}-Gomis, Mikel Forcada, Amir Kamran, Faheem Kirefu, Philipp
  Koehn, Sergio Ortiz-Rojas, Leopoldo Pla, Gema Ram{\'i}rez-S{\'a}nchez, Elsa
  Sarr{\'i}as, Marek Strelec, Brian Thompson, William Waites, Dion Wiggins, and
  Jaume Zaragoza.
\newblock Paracrawl: Web-scale acquisition of parallel corpora.
\newblock In \emph{Proceedings of the 58th Annual Meeting of the Association
  for Computational Linguistics}, page 4555–4567, July 2020.
\newblock \doi{10.18653/v1/2020.acl-main.417}.
\newblock URL \url{https://acl2020.org/}.

\bibitem[Bhattacharya et~al.(2019)Bhattacharya, Chowdhury, and
  Raykar]{bhattacharya2019multimodal}
Indrani Bhattacharya, Arkabandhu Chowdhury, and Vikas~C Raykar.
\newblock Multimodal dialog for browsing large visual catalogs using
  exploration-exploitation paradigm in a joint embedding space.
\newblock In \emph{Proceedings of the 2019 on International Conference on
  Multimedia Retrieval}, pages 187--191, 2019.

\bibitem[Bojar et~al.(2014)Bojar, Buck, Federmann, Haddow, Koehn, Leveling,
  Monz, Pecina, Post, Saint-Amand, Soricut, Specia, and
  Tamchyna]{bojar-EtAl:2014:W14-33}
Ondrej Bojar, Christian Buck, Christian Federmann, Barry Haddow, Philipp Koehn,
  Johannes Leveling, Christof Monz, Pavel Pecina, Matt Post, Herve Saint-Amand,
  Radu Soricut, Lucia Specia, and Ale~{s} Tamchyna.
\newblock Findings of the 2014 workshop on statistical machine translation.
\newblock In \emph{Proceedings of the Ninth Workshop on Statistical Machine
  Translation}, pages 12--58, June 2014.
\newblock URL \url{http://www.aclweb.org/anthology/W/W14/W14-3302}.

\bibitem[Chung et~al.(2014)Chung, Gulcehre, Cho, and
  Bengio]{chung2014empirical}
Junyoung Chung, Caglar Gulcehre, KyungHyun Cho, and Yoshua Bengio.
\newblock Empirical evaluation of gated recurrent neural networks on sequence
  modeling.
\newblock \emph{arXiv preprint arXiv:1412.3555}, 2014.

\bibitem[Das et~al.(2017)Das, Kottur, Gupta, Singh, Yadav, Moura, Parikh, and
  Batra]{das2017visual}
Abhishek Das, Satwik Kottur, Khushi Gupta, Avi Singh, Deshraj Yadav,
  Jos{\'e}~MF Moura, Devi Parikh, and Dhruv Batra.
\newblock Visual dialog.
\newblock In \emph{Proceedings of the IEEE Conference on Computer Vision and
  Pattern Recognition}, pages 326--335, 2017.

\bibitem[Devlin et~al.(2018)Devlin, Chang, Lee, and Toutanova]{devlin2018bert}
Jacob Devlin, Ming-Wei Chang, Kenton Lee, and Kristina Toutanova.
\newblock Bert: Pre-training of deep bidirectional transformers for language
  understanding.
\newblock \emph{arXiv preprint arXiv:1810.04805}, 2018.

\bibitem[Doersch and Zisserman(2017)]{doersch2017multi}
Carl Doersch and Andrew Zisserman.
\newblock Multi-task self-supervised visual learning.
\newblock In \emph{Proceedings of the IEEE International Conference on Computer
  Vision}, pages 2051--2060, 2017.

\bibitem[Doersch et~al.(2015)Doersch, Gupta, and
  Efros]{doersch2015unsupervised}
Carl Doersch, Abhinav Gupta, and Alexei~A Efros.
\newblock Unsupervised visual representation learning by context prediction.
\newblock In \emph{Proceedings of the IEEE International Conference on Computer
  Vision}, pages 1422--1430, 2015.

\bibitem[Dong et~al.(2015)Dong, Loy, He, and Tang]{dong2015image}
Chao Dong, Chen~Change Loy, Kaiming He, and Xiaoou Tang.
\newblock Image super-resolution using deep convolutional networks.
\newblock \emph{IEEE transactions on pattern analysis and machine
  intelligence}, 38\penalty0 (2):\penalty0 295--307, 2015.

\bibitem[Dosovitskiy et~al.(2015)Dosovitskiy, Fischer, Springenberg,
  Riedmiller, and Brox]{dosovitskiy2015discriminative}
Alexey Dosovitskiy, Philipp Fischer, Jost~Tobias Springenberg, Martin
  Riedmiller, and Thomas Brox.
\newblock Discriminative unsupervised feature learning with exemplar
  convolutional neural networks.
\newblock \emph{IEEE transactions on pattern analysis and machine
  intelligence}, 38\penalty0 (9):\penalty0 1734--1747, 2015.

\bibitem[Gidaris et~al.(2018)Gidaris, Singh, and
  Komodakis]{gidaris2018unsupervised}
Spyros Gidaris, Praveer Singh, and Nikos Komodakis.
\newblock Unsupervised representation learning by predicting image rotations.
\newblock \emph{arXiv preprint arXiv:1803.07728}, 2018.

\bibitem[Girshick et~al.(2014)Girshick, Donahue, Darrell, and
  Malik]{girshick2014rich}
Ross Girshick, Jeff Donahue, Trevor Darrell, and Jitendra Malik.
\newblock Rich feature hierarchies for accurate object detection and semantic
  segmentation.
\newblock In \emph{Proceedings of the IEEE conference on computer vision and
  pattern recognition}, pages 580--587, 2014.

\bibitem[Goodfellow et~al.(2014)Goodfellow, Pouget-Abadie, Mirza, Xu,
  Warde-Farley, Ozair, Courville, and Bengio]{GAN}
Ian Goodfellow, Jean Pouget-Abadie, Mehdi Mirza, Bing Xu, David Warde-Farley,
  Sherjil Ozair, Aaron Courville, and Yoshua Bengio.
\newblock Generative adversarial nets.
\newblock In \emph{Conference on Neural Information Processing Systems}, 2014.

\bibitem[Graves et~al.(2013)Graves, Mohamed, and Hinton]{graves2013speech}
Alex Graves, Abdel-rahman Mohamed, and Geoffrey Hinton.
\newblock Speech recognition with deep recurrent neural networks.
\newblock In \emph{2013 IEEE international conference on acoustics, speech and
  signal processing}, pages 6645--6649. IEEE, 2013.

\bibitem[Guo et~al.(2019)Guo, Wu, Gao, Rennie, and Feris]{guo2019fashion}
Xiaoxiao Guo, Hui Wu, Yupeng Gao, Steven Rennie, and Rogerio Feris.
\newblock Fashion iq: A new dataset towards retrieving images by natural
  language feedback.
\newblock \emph{arXiv preprint arXiv:1905.12794}, 2019.

\bibitem[Hjelm et~al.(2018)Hjelm, Fedorov, Lavoie-Marchildon, Grewal, Bachman,
  Trischler, and Bengio]{hjelm2018learning}
R~Devon Hjelm, Alex Fedorov, Samuel Lavoie-Marchildon, Karan Grewal, Phil
  Bachman, Adam Trischler, and Yoshua Bengio.
\newblock Learning deep representations by mutual information estimation and
  maximization.
\newblock \emph{arXiv preprint arXiv:1808.06670}, 2018.

\bibitem[Hou et~al.(2020)Hou, Suominen, Koniusz, Caldwell, and
  Gedeon]{token_cnn}
Weiwei Hou, Hanna Suominen, Piotr Koniusz, Sabrina~B. Caldwell, and Tom Gedeon.
\newblock A token-wise cnn-based method for sentence compression.
\newblock In \emph{Neural Information Processing - 27th International
  Conference, {ICONIP}}, volume 12532 of \emph{Lecture Notes in Computer
  Science}, pages 668--679. Springer, 2020.
\newblock \doi{10.1007/978-3-030-63830-6\_56}.

\bibitem[Johnson et~al.(2019)Johnson, Douze, and J{\'e}gou]{johnson2019billion}
Jeff Johnson, Matthijs Douze, and Herv{\'e} J{\'e}gou.
\newblock Billion-scale similarity search with gpus.
\newblock \emph{IEEE Transactions on Big Data}, 2019.

\bibitem[Koniusz and Zhang(2020)]{kon_tpami2020a}
Piotr Koniusz and Hongguang Zhang.
\newblock Power normalizations in fine-grained image, few-shot image and graph
  classification.
\newblock In \emph{IEEE Transactions on Pattern Analysis and Machine
  Intelligence}. IEEE, 2020.
\newblock \doi{10.1109/TPAMI.2021.3107164}.

\bibitem[Koniusz et~al.(2013)Koniusz, Yan, and Mikolajczyk]{pk_cviu}
Piotr Koniusz, Fei Yan, and Krystian Mikolajczyk.
\newblock Comparison of mid-level feature coding approaches and pooling
  strategies in visual concept detection.
\newblock \emph{Computer Vision and Image Understanding}, 117\penalty0
  (5):\penalty0 479–492, May 2013.
\newblock ISSN 1077-3142.
\newblock \doi{10.1016/j.cviu.2012.10.010}.
\newblock URL \url{https://doi.org/10.1016/j.cviu.2012.10.010}.

\bibitem[Koniusz et~al.(2017{\natexlab{a}})Koniusz, Tas, and
  Porikli]{koniusz2017domain}
Piotr Koniusz, Yusuf Tas, and Fatih Porikli.
\newblock Domain adaptation by mixture of alignments of second-or higher-order
  scatter tensors.
\newblock \emph{Proceedings of the IEEE Conference on Computer Vision and
  Pattern Recognition}, 2, 2017{\natexlab{a}}.

\bibitem[Koniusz et~al.(2017{\natexlab{b}})Koniusz, Yan, Gosselin, and
  Mikolajczyk]{pk_hop}
Piotr Koniusz, Fei Yan, Philippe-Henri Gosselin, and Krystian Mikolajczyk.
\newblock Higher-order occurrence pooling for bags-of-words: Visual concept
  detection.
\newblock \emph{IEEE Transactions on Pattern Analysis and Machine
  Intelligence}, 39\penalty0 (2):\penalty0 313--326, 2017{\natexlab{b}}.
\newblock \doi{10.1109/TPAMI.2016.2545667}.

\bibitem[Koniusz et~al.(2018)Koniusz, Tas, Zhang, Harandi, Porikli, and
  Zhang]{Koniusz2018Museum}
Piotr Koniusz, Yusuf Tas, Hongguang Zhang, Mehrtash Harandi, Fatih Porikli, and
  Rui Zhang.
\newblock Museum exhibit identification challenge for the supervised domain
  adaptation and beyond.
\newblock In \emph{The European Conference on Computer Vision}, 2018.

\bibitem[Kottur et~al.(2019)Kottur, Moura, Parikh, Batra, and
  Rohrbach]{kottur2019clevr}
Satwik Kottur, Jos{\'e}~MF Moura, Devi Parikh, Dhruv Batra, and Marcus
  Rohrbach.
\newblock Clevr-dialog: A diagnostic dataset for multi-round reasoning in
  visual dialog.
\newblock \emph{arXiv preprint arXiv:1903.03166}, 2019.

\bibitem[Liu et~al.(2016)Liu, Luo, Qiu, Wang, and Tang]{liu2016deepfashion}
Ziwei Liu, Ping Luo, Shi Qiu, Xiaogang Wang, and Xiaoou Tang.
\newblock Deepfashion: Powering robust clothes recognition and retrieval with
  rich annotations.
\newblock In \emph{Proceedings of the IEEE conference on computer vision and
  pattern recognition}, pages 1096--1104, 2016.

\bibitem[Marian and Shook(2012)]{cog_ben}
Viorica Marian and Anthony Shook.
\newblock The cognitive benefits of being bilingual.
\newblock \emph{Cerebrum : the Dana forum on brain science}, 2012:\penalty0 13,
  10 2012.

\bibitem[Mikolov et~al.(2010)Mikolov, Karafi{\'a}t, Burget, {\v{C}}ernock{\`y},
  and Khudanpur]{mikolov2010recurrent}
Tom{\'a}{\v{s}} Mikolov, Martin Karafi{\'a}t, Luk{\'a}{\v{s}} Burget, Jan
  {\v{C}}ernock{\`y}, and Sanjeev Khudanpur.
\newblock Recurrent neural network based language model.
\newblock In \emph{Eleventh annual conference of the international speech
  communication association}, 2010.

\bibitem[Moon et~al.(2020)Moon, Kottur, Crook, De, Poddar, Levin, Whitney,
  Difranco, Beirami, Cho, Subba, and Geramifard]{crook2020simmc}
Seungwhan Moon, Satwik Kottur, Paul~A. Crook, Ankita De, Shivani Poddar,
  Theodore Levin, David Whitney, Daniel Difranco, Ahmad Beirami, Eunjoon Cho,
  Rajen Subba, and Alborz Geramifard.
\newblock Situated and interactive multimodal conversations, 2020.

\bibitem[Ott et~al.(2018)Ott, Edunov, Grangier, and Auli]{ott2018scaling}
Myle Ott, Sergey Edunov, David Grangier, and Michael Auli.
\newblock Scaling neural machine translation.
\newblock In \emph{Proceedings of the Third Conference on Machine Translation
  (WMT)}, 2018.

\bibitem[Ram et~al.(2018)Ram, Prasad, Khatri, Venkatesh, Gabriel, Liu, Nunn,
  Hedayatnia, Cheng, Nagar, et~al.]{ram2018conversational}
Ashwin Ram, Rohit Prasad, Chandra Khatri, Anu Venkatesh, Raefer Gabriel, Qing
  Liu, Jeff Nunn, Behnam Hedayatnia, Ming Cheng, Ashish Nagar, et~al.
\newblock Conversational ai: The science behind the alexa prize.
\newblock \emph{arXiv preprint arXiv:1801.03604}, 2018.

\bibitem[Ritter et~al.(2010)Ritter, Cherry, and Dolan]{ritter2010unsupervised}
Alan Ritter, Colin Cherry, and Bill Dolan.
\newblock Unsupervised modeling of twitter conversations.
\newblock In \emph{Human Language Technologies: The 2010 Annual Conference of
  the North American Chapter of the Association for Computational Linguistics},
  pages 172--180, 2010.

\bibitem[Saha et~al.(2018)Saha, Khapra, and Sankaranarayanan]{saha2018towards}
Amrita Saha, Mitesh~M Khapra, and Karthik Sankaranarayanan.
\newblock Towards building large scale multimodal domain-aware conversation
  systems.
\newblock In \emph{Thirty-Second AAAI Conference on Artificial Intelligence},
  2018.

\bibitem[Serban et~al.(2015)Serban, Sordoni, Bengio, Courville, and
  Pineau]{serban2015building}
Iulian~V Serban, Alessandro Sordoni, Yoshua Bengio, Aaron Courville, and Joelle
  Pineau.
\newblock Building end-to-end dialogue systems using generative hierarchical
  neural network models.
\newblock \emph{arXiv preprint arXiv:1507.04808}, 2015.

\bibitem[Serban et~al.(2016)Serban, Sordoni, Bengio, Courville, and
  Pineau]{serban2016building}
Iulian~V Serban, Alessandro Sordoni, Yoshua Bengio, Aaron Courville, and Joelle
  Pineau.
\newblock Building end-to-end dialogue systems using generative hierarchical
  neural network models.
\newblock In \emph{Thirtieth AAAI Conference on Artificial Intelligence}, 2016.

\bibitem[Serban et~al.(2017)Serban, Sordoni, Lowe, Charlin, Pineau, Courville,
  and Bengio]{serban2017hierarchical}
Iulian~Vlad Serban, Alessandro Sordoni, Ryan Lowe, Laurent Charlin, Joelle
  Pineau, Aaron Courville, and Yoshua Bengio.
\newblock A hierarchical latent variable encoder-decoder model for generating
  dialogues.
\newblock In \emph{Thirty-First AAAI Conference on Artificial Intelligence},
  2017.

\bibitem[Shang et~al.(2015)Shang, Lu, and Li]{shang2015neural}
Lifeng Shang, Zhengdong Lu, and Hang Li.
\newblock Neural responding machine for short-text conversation.
\newblock \emph{arXiv preprint arXiv:1503.02364}, 2015.

\bibitem[Shiri et~al.(2019{\natexlab{a}})Shiri, Yu, Porikli, Hartley, and
  Koniusz]{ShiriYPHK19}
Fatemeh Shiri, Xin Yu, Fatih Porikli, Richard Hartley, and Piotr Koniusz.
\newblock Recovering faces from portraits with auxiliary facial attributes.
\newblock In \emph{Winter Conference on Applications of Computer Vision}, pages
  406--415, 2019{\natexlab{a}}.

\bibitem[Shiri et~al.(2019{\natexlab{b}})Shiri, Yu, Porikli, Hartley, and
  Koniusz]{fatima_ijcv}
Fatemeh Shiri, Xin Yu, Fatih Porikli, Richard Hartley, and Piotr Koniusz.
\newblock Identity-preserving face recovery from stylized portraits.
\newblock \emph{International Journal of Computer Vision}, 127\penalty0
  (6-7):\penalty0 863--883, 2019{\natexlab{b}}.
\newblock \doi{10.1007/s11263-019-01169-1}.

\bibitem[Simonyan and Zisserman(2014)]{simonyan2014very}
Karen Simonyan and Andrew Zisserman.
\newblock Very deep convolutional networks for large-scale image recognition.
\newblock \emph{arXiv preprint arXiv:1409.1556}, 2014.

\bibitem[Sordoni et~al.(2015)Sordoni, Galley, Auli, Brockett, Ji, Mitchell,
  Nie, Gao, and Dolan]{sordoni2015neural}
Alessandro Sordoni, Michel Galley, Michael Auli, Chris Brockett, Yangfeng Ji,
  Margaret Mitchell, Jian-Yun Nie, Jianfeng Gao, and Bill Dolan.
\newblock A neural network approach to context-sensitive generation of
  conversational responses.
\newblock \emph{arXiv preprint arXiv:1506.06714}, 2015.

\bibitem[Tas and Koniusz(2018)]{action_da}
Y.~Tas and P.~Koniusz.
\newblock Cnn-based action recognition and supervised domain adaptation on 3d
  body skeletons via kernel feature maps.
\newblock \emph{British Machine Vision Conference}, 2018.

\bibitem[Thomason et~al.(2020)Thomason, Murray, Cakmak, and
  Zettlemoyer]{thomason2020vision}
Jesse Thomason, Michael Murray, Maya Cakmak, and Luke Zettlemoyer.
\newblock Vision-and-dialog navigation.
\newblock In \emph{Conference on Robot Learning}, pages 394--406, 2020.

\bibitem[Vaswani et~al.(2017)Vaswani, Shazeer, Parmar, Uszkoreit, Jones, Gomez,
  Kaiser, and Polosukhin]{46201}
Ashish Vaswani, Noam Shazeer, Niki Parmar, Jakob Uszkoreit, Llion Jones,
  Aidan~N Gomez, \L~ukasz Kaiser, and Illia Polosukhin.
\newblock Attention is all you need.
\newblock In \emph{Advances in Neural Information Processing Systems},
  volume~30. Curran Associates, Inc., 2017.

\bibitem[Wang and Koniusz(2021)]{i3d_halluc2}
Lei Wang and Piotr Koniusz.
\newblock Hallucinating statistical moment and subspace descriptors for action
  recognition.
\newblock \emph{ACM Multimedia}, 2021.

\bibitem[Wang et~al.(2019)Wang, Koniusz, and Huynh]{i3d_halluc}
Lei Wang, Piotr Koniusz, and Du~Q. Huynh.
\newblock Hallucinating idt descriptors and i3d optical flow features for
  action recognition with cnns.
\newblock \emph{International Conference on Computer Vision}, 2019.

\bibitem[Wen et~al.(2016)Wen, Vandyke, Mrksic, Gasic, Rojas-Barahona, Su,
  Ultes, and Young]{wen2016network}
Tsung-Hsien Wen, David Vandyke, Nikola Mrksic, Milica Gasic, Lina~M
  Rojas-Barahona, Pei-Hao Su, Stefan Ultes, and Steve Young.
\newblock A network-based end-to-end trainable task-oriented dialogue system.
\newblock \emph{arXiv preprint arXiv:1604.04562}, 2016.

\bibitem[Woodworth and Thorndike(1901)]{woodworth_particle}
R.~S. Woodworth and E.~L. Thorndike.
\newblock The influence of improvement in one mental function upon the
  efficiency of other functions.
\newblock \emph{Psychological Review (I)}, 8\penalty0 (3):\penalty0 247--261,
  1901.
\newblock \doi{10.1037/h0074898}.

\bibitem[Wu et~al.(2018)Wu, Hu, and Mooney]{wu2018joint}
Jialin Wu, Zeyuan Hu, and Raymond~J Mooney.
\newblock Joint image captioning and question answering.
\newblock \emph{arXiv preprint arXiv:1805.08389}, 2018.

\bibitem[Xu et~al.(2015)Xu, Ba, Kiros, Cho, Courville, Salakhudinov, Zemel, and
  Bengio]{xu2015show}
Kelvin Xu, Jimmy Ba, Ryan Kiros, Kyunghyun Cho, Aaron Courville, Ruslan
  Salakhudinov, Rich Zemel, and Yoshua Bengio.
\newblock Show, attend and tell: Neural image caption generation with visual
  attention.
\newblock In \emph{International conference on machine learning}, pages
  2048--2057, 2015.

\bibitem[Zeng et~al.(2017)Zeng, Chen, Chuang, Liao, Niebles, and
  Sun]{zeng2017leveraging}
Kuo-Hao Zeng, Tseng-Hung Chen, Ching-Yao Chuang, Yuan-Hong Liao, Juan~Carlos
  Niebles, and Min Sun.
\newblock Leveraging video descriptions to learn video question answering.
\newblock In \emph{Thirty-First AAAI Conference on Artificial Intelligence},
  2017.

\bibitem[Zhang et~al.(2020{\natexlab{a}})Zhang, Zhang, Qi, Li, Torr, and
  Koniusz]{zhang2020few}
H~Zhang, L~Zhang, X~Qi, H~Li, PHS Torr, and P~Koniusz.
\newblock Few-shot action recognition with permutation-invariant attention.
\newblock In \emph{The European Conference on Computer Vision},
  2020{\natexlab{a}}.

\bibitem[Zhang and Koniusz(2019)]{sosn}
Hongguang Zhang and Piotr Koniusz.
\newblock Power normalizing second-order similarity network for few-shot
  learning.
\newblock \emph{Winter Conference on Applications of Computer Vision}, 2019.

\bibitem[Zhang et~al.(2021)Zhang, Koniusz, Jian, Li, and Torr]{arl}
Hongguang Zhang, Piotr Koniusz, Songlei Jian, Hongdong Li, and Philip H.~S.
  Torr.
\newblock Rethinking class relations: Absolute-relative supervised and
  unsupervised few-shot learning.
\newblock In \emph{Proceedings of the IEEE Conference on Computer Vision and
  Pattern Recognition}, pages 9432--9441, 2021.

\bibitem[Zhang et~al.(2017)Zhang, Tas, and Koniusz]{zhang2018artwork}
Rui Zhang, Yusuf Tas, and Piotr Koniusz.
\newblock Artwork identification from wearable camera images for enhancing
  experience of museum audiences.
\newblock In \emph{Museums and the Web}, 2017.

\bibitem[Zhang et~al.(2020{\natexlab{b}})Zhang, Luo, Wang, and
  Koniusz]{Zhang_2020_ACCV}
Shan Zhang, Dawei Luo, Lei Wang, and Piotr Koniusz.
\newblock Few-shot object detection by second-order pooling.
\newblock \emph{Asian Conference on Computer Vision}, 2020{\natexlab{b}}.

\bibitem[Zhu and Koniusz(2021{\natexlab{a}})]{refine}
Hao Zhu and Piotr Koniusz.
\newblock {REFINE: Random RangE FInder for Network Embedding}.
\newblock In \emph{{ACM International Conference on Information and Knowledge
  Management}}, 2021{\natexlab{a}}.
\newblock \doi{0.1145/3459637.3482168}.

\bibitem[Zhu and Koniusz(2021{\natexlab{b}})]{ssgc}
Hao Zhu and Piotr Koniusz.
\newblock Simple spectral graph convolution.
\newblock In \emph{International Conference on Learning Representations},
  2021{\natexlab{b}}.

\bibitem[Zhu et~al.(2021)Zhu, Sun, and Koniusz]{coles}
Hao Zhu, Ke~Sun, and Piotr Koniusz.
\newblock Contrastive laplacian eigenmaps.
\newblock In \emph{Conference on Neural Information Processing Systems}, 2021.

\end{thebibliography}
\end{document}
}